\documentclass[sigconf]{acmart} %显示作者

\usepackage{color}
\usepackage{makecell}
\usepackage{multirow}
\usepackage{amsthm}
\newtheorem{theorem}{Theorem}
\newtheorem{definition}{Definition}
\newtheorem{corollary}{Corollary}

\usepackage{subfigure}
\usepackage{amsmath,amsfonts}
\usepackage[ruled,vlined,linesnumbered,ruled]{algorithm2e}

\usepackage{graphicx}

\AtBeginDocument{%
  \providecommand\BibTeX{{%
    \normalfont B\kern-0.5em{\scshape i\kern-0.25em b}\kern-0.8em\TeX}}}

\copyrightyear{2024}
\acmYear{2024}
\setcopyright{acmlicensed}
\acmConference[KDD '24]{Proceedings of the 30th ACM SIGKDD Conference on Knowledge Discovery and Data Mining}{August 25--29, 2024}{Barcelona, Spain}
\acmBooktitle{Proceedings of the 30th ACM SIGKDD Conference on Knowledge Discovery and Data Mining (KDD '24), August 25--29, 2024, Barcelona, Spain}
\acmDOI{10.1145/3637528.3671704}
\acmISBN{979-8-4007-0490-1/24/08}

\begin{document}

%%
%% The "title" command has an optional parameter,
%% allowing the author to define a "short title" to be used in page headers.
%\title{An Unsupervised Learning Framework Combined with Heuristics for the Maximum Minimal Cut Problem}

\title{An Unsupervised Learning Framework Combined with Heuristics for the Maximum Minimal Cut Problem}

%%
%% The "author" command and its associated commands are used to define
%% the authors and their affiliations.
%% Of note is the shared affiliation of the first two authors, and the
%% "authornote" and "authornotemark" commands
%% used to denote shared contribution to the research.
\author{Huaiyuan Liu}
\affiliation{%
  \institution{Harbin Institute of Technology}
  \city{Harbin}
  \country{China}}
\email{hyliu@hit.edu.cn}

\author{Xianzhang Liu}
\affiliation{%
  \institution{Harbin Institute of Technology}
  \city{Harbin}
  \country{China}}
\email{liuxz_hit@foxmail.com}

\author{Donghua Yang}
\affiliation{%
  \institution{Harbin Institute of Technology}
  \city{Harbin}
  \country{China}}
\email{yang.dh@hit.edu.cn}

\author{Hongzhi Wang}
%\authornotemark[1]
\authornote{Hongzhi Wang is the corresponding author.}
\affiliation{%
  \institution{Harbin Institute of Technology}
  \city{Harbin}
  \country{China}}
\email{wangzh@hit.edu.cn}

\author{Yinchi Long}
\affiliation{%
  \institution{Harbin Institute of Technology}
  \city{Harbin}
  \country{China}}
\email{i@lyc.dev}

\author{Mengtong Ji}
\affiliation{%
  \institution{Harbin Institute of Technology}
  \city{Harbin}
  \country{China}}
\email{2512508310@qq.com}

\author{Dongjing Miao}
\affiliation{%
  \institution{Harbin Institute of Technology}
  \city{Harbin}
  \country{China}}
\email{miaodongjing@hit.edu.cn}

\author{Zhiyu Liang}
\affiliation{%
  \institution{Harbin Institute of Technology}
  \city{Harbin}
  \country{China}}
\email{zyliang@hit.edu.cn}

%%
%% By default, the full list of authors will be used in the page
%% headers. Often, this list is too long, and will overlap
%% other information printed in the page headers. This command allows
%% the author to define a more concise list
%% of authors' names for this purpose.
\renewcommand{\shortauthors}{Huaiyuan Liu et al.}

%%
%% The abstract is a short summary of the work to be presented in the
%% article.
\begin{abstract}
The Maximum Minimal Cut Problem (MMCP), a NP-hard combinatorial optimization (CO) problem, has not received much attention due to the demanding and challenging bi-connectivity constraint. Moreover, as a CO problem, it is also a daunting task for machine learning, especially without labeled instances. To deal with these problems, this work proposes an unsupervised learning framework combined with heuristics for MMCP that can provide valid and high-quality solutions. As far as we know, this is the first work that explores machine learning and heuristics to solve MMCP. The unsupervised solver is inspired by a relaxation-plus-rounding approach, the relaxed solution is parameterized by graph neural networks, and the cost and penalty of MMCP are explicitly written out, which can train the model end-to-end. A crucial observation is that each solution corresponds to at least one spanning tree. Based on this finding, a heuristic solver that implements tree transformations by adding vertices is utilized to repair and improve the solution quality of the unsupervised solver. Alternatively, the graph is simplified while guaranteeing solution consistency, which reduces the running time. We conduct extensive experiments to evaluate our framework and give a specific application. The results demonstrate the superiority of our method against two techniques designed. 
\end{abstract}

%%
%% The code below is generated by the tool at http://dl.acm.org/ccs.cfm.
%% Please copy and paste the code instead of the example below.
%%
\begin{CCSXML}
<ccs2012>
   <concept>
       <concept_id>10002950.10003624.10003625.10003630</concept_id>
       <concept_desc>Mathematics of computing~Combinatorial optimization</concept_desc>
       <concept_significance>500</concept_significance>
       </concept>
   <concept>
       <concept_id>10010147.10010257.10010258.10010260</concept_id>
       <concept_desc>Computing methodologies~Unsupervised learning</concept_desc>
       <concept_significance>500</concept_significance>
       </concept>
 </ccs2012>
\end{CCSXML}

\ccsdesc[500]{Mathematics of computing~Combinatorial optimization}
\ccsdesc[500]{Computing methodologies~Unsupervised learning}
% \ccsdesc[500]{Computing methodologies~Heuristic function construction}

%%
%% Keywords. The author(s) should pick words that accurately describe
%% the work being presented. Separate the keywords with commas.
\keywords{combinatorial optimization, maximum minimal cut problem, unsupervised learning, heuristics}
%% A "teaser" image appears between the author and affiliation
%% information and the body of the document, and typically spans the
%% page.
%\begin{teaserfigure}
%  \includegraphics[width=\textwidth]{figure0}
%  \caption{Seattle Mariners at Spring Training, 2010.}
%  \Description{Enjoying the baseball game from the third-base
%  seats. Ichiro Suzuki preparing to bat.}
%  \label{fig:teaser}
%\end{teaserfigure}

\received{08 February 2024}
\received[revised]{18 April 2024}
\received[accepted]{16 May 2024}

%%
%% This command processes the author and affiliation and title
%% information and builds the first part of the formatted document.
\maketitle

\section{Introduction}

The graph is a fundamental structure used to depict diverse relationships among entities \cite{ge2023efficient,sun2022efficient}, such as social networks, communication networks, biological information, and power grids. Combinatorial optimization (CO) on graphs is a crucial field in computational mathematics, with typical examples including the Max-cut \cite{Garey79}, Maximum Independent Set (MIS) \cite{Miller60}, and Traveling Salesman Problem (TSP) \cite{Voigt31} applied in various scenarios. The exact solutions for most combinatorial optimization problems are intractable to search, primarily due to their NP-hard nature. Consequently, significant research efforts have been dedicated to solving these tasks by generating approximate solutions \cite{Goemans94,Lin73}. 

\begin{figure}[ht]
\centering    %居中
\subfigure[Max-cut.] %第一张子图
{\begin{minipage}{0.32\columnwidth}
\label{fig:1-1}
\centering          %子图居中
\includegraphics[scale=0.45]{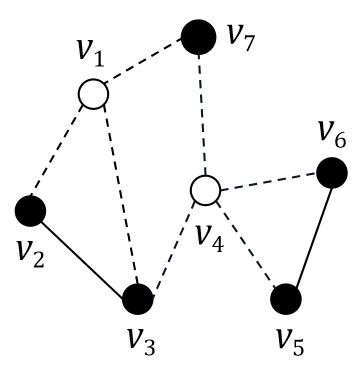}   %以pic.jpg的0.5倍大小输出
\end{minipage}}	
\subfigure[MMC.] %第二张子图
{\begin{minipage}{0.32\columnwidth}
\label{fig:1-2}
\centering      %子图居中
\includegraphics[scale=0.45]{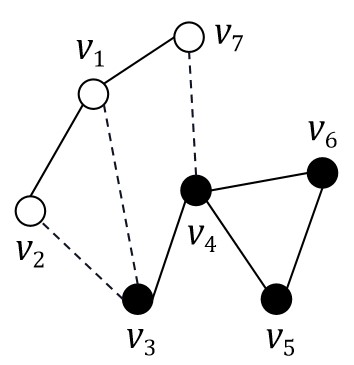}   %以pic.jpg的0.5倍大小输出
\end{minipage}}
\subfigure[CMC.] %第三张子图
{\begin{minipage}{0.32\columnwidth}
\label{fig:1-3}
\centering      %子图居中
\includegraphics[scale=0.45]{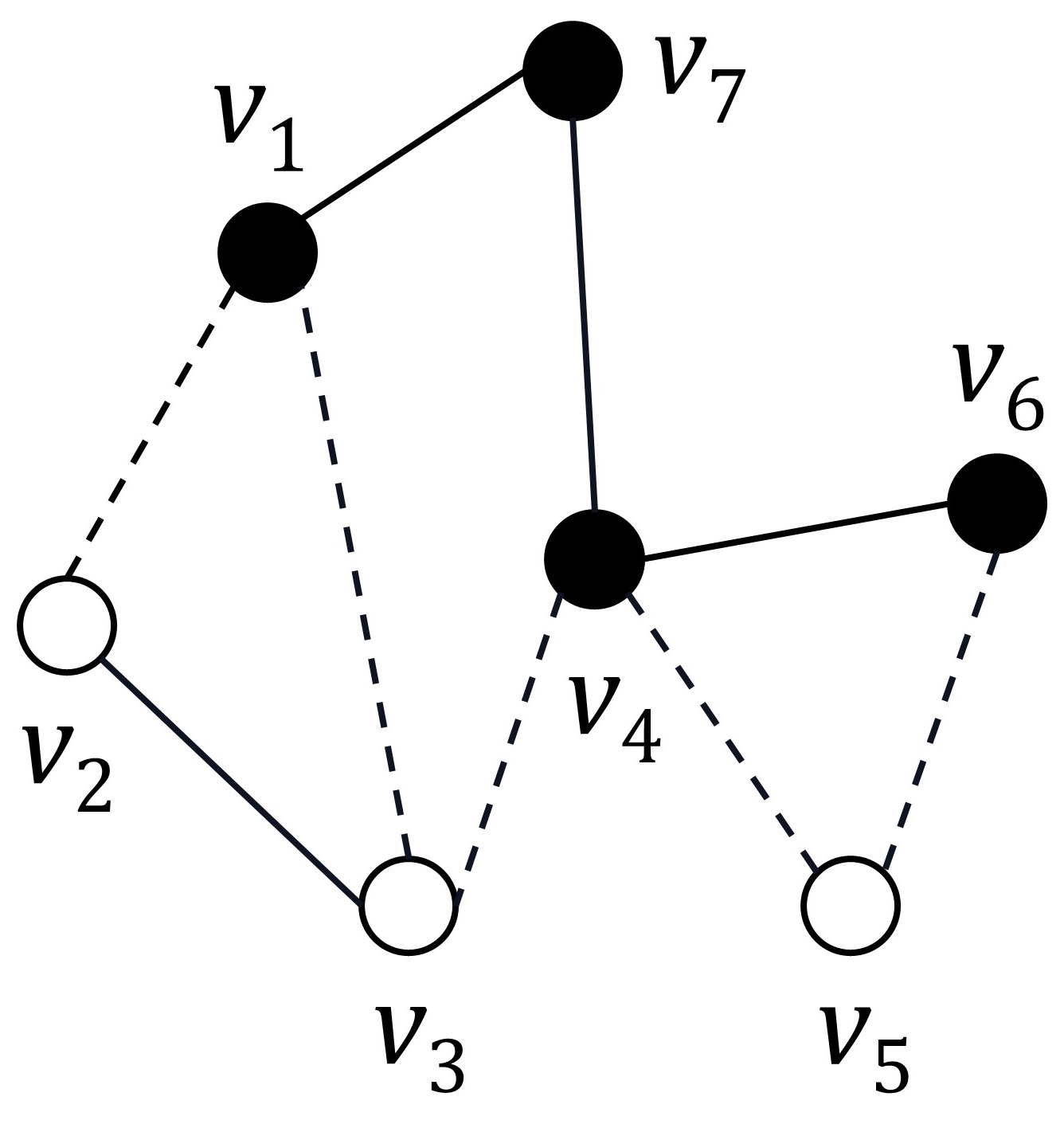}   %以pic.jpg的0.5倍大小输出
\end{minipage}}
\caption{An example to illustrate the difference among max-cut, maximum minimal cut, and connected maximum cut.} %  %大图名称
\label{fig:1}  %图片引用标记
\end{figure}

The \emph{Max-cut} problem, a classic NP-hard problem in graph theory, has been extensively studied for decades, while the \emph{Maximum Minimal Cut} (MMC) problem as a variant of the max-cut problem is rarely mentioned. Given a connected graph $G = (V, E)$, the objective of the max-cut problem is to discover a partition method that divides all vertices into two complementary sets, denoted as $K$ and $L$, in a way that maximizes the cardinality of edges between $K$ and $L$. The maximum minimal cut problem (MMCP) adds the requirement that both $K$ and $L$ be connected based on the max-cut problem, also called the largest bond problem \cite{Duarte21}. As a well-known fact, a cut $C=(K, L)$ of $G$ is minimal if and only if both subgraphs induced by $K$ and $L$ are connected~\cite{Diestel17}, where $K, L \in V$ and $L=V\backslash K$. Therefore, a minimal cut is regarded as a two-sided connected cut~\cite{Eto19}. Additionally, the Connected Maximum Cut (CMC) problem is a variant of the max-cut problem that only demands the connectivity of $K$. For example, in Figure~\ref{fig:1}, $\{ v_1, v_4\}$, $\{v_1, v_2, v_7\}$ and $\{v_1, v_4, v_6, v_7\}$ are max-cut, maximum minimal cut, and connected maximum cut of $G$, respectively. Note that the notable distinction among the three problems lies in the connectivity of the two subsets after separation.

The Max-cut, as a fundamental problem of CO, can find applications in image segmentation~\cite{Dunning18, Lange19}, network analysis~\cite{Barahona96}, and statistical physics~\cite{Barahona88}, etc., while the MMCP rarely applied in real-world scenarios. Recently, we noticed the significant importance of MMCP as a mathematical form of computation in identifying key cross-sections for power grids \cite{zhu2021comprehensive}, and we left the specifics for Section \ref{subsec:case}. Inspired by this, the MMCP will play an indispensable role in the monitoring and early warning of related complex networks. Over the past decades, researchers have made significant efforts to prove the complexity and design parameterized algorithms for this issue \cite{Duarte21, Flynn17}. However, no fast solver for MMCP has been released so far, which is challenging due to the following reasons:

\emph{Special Constraint.} Unsupervised learning is widely used as a popular paradigm for combinatorial optimization problems~\cite{yao2019experimental, amizadeh2018learning, WangL23} that obtains solutions by minimizing a differentiable loss function, whose successful training hinges on empirically-identified correction terms \cite{karalias2020erdos, amizadeh2019pdp}. However, no methods can effectively represent the bi-connectivity constraints of the MMCP in an algebraic form and cannot be trained through back-propagation. Alternatively, it is exceptionally difficult to efficiently decode the relaxed solution of a neural network to a discrete solution that satisfies the constraints \cite{li2018combinatorial}, especially without a fully labeled solution.

\emph{Lack of Equivalence.} We claim that designing solvers for MMCP is often of greater difficulty than traditional CO problems because a general solver cannot be implemented for MMCP due to the lack of closed-form equivalence of solutions. In other words, compressing the solution space to a solution space that satisfies the constraints is a formidable task. No study can equate the solutions of MMCP to the solutions that disregard the constraints. For heuristics, it is more efficient to explore a larger objective value in constraint-satisfying solutions than to judge the legitimacy after obtaining the solutions.

\emph{Weak Universality.} Connectivity is a feature strongly associated with the graph structure, leading to the large obstacle of solving MMCP uniformly for the graphs with different structures. Regrettably, most existing approaches to address MMCP are based on the assumptions of graph structures, e.g., planar graph and simple 3-connected graph \cite{Ding16, Flynn17}. Only graphs that satisfy specific structure constraints are considered. Therefore, overcoming the effects of graph structures to construct a unified framework is crucial.

% Thus, the known results of MMCP are few due to the typical intractability of simultaneously maximizing the goal and ensuring the connectivity of cuts.

% Our framework is dedicated to overcoming some of the aforementioned obstacles of

\textbf{Contributions.} To tackle the aforementioned issues and fill the gap, we design a novel \emph{unsu\textbf{\underline{p}}ervised learn\textbf{\underline{i}}ng framew\textbf{\underline{o}}rk combi\textbf{\underline{ne}}d with h\textbf{\underline{e}}u\textbf{\underline{r}}istics (PIONEER)} and provide new ideas for solving the maximum minimal cut problem in this work. In summary, the main contributions of our work are summarized as follows:

\begin{itemize}
\item {\textit{Pioneering.}} We propose PIONEER, to the best of our knowledge, this is the first study on the unsupervised learning and
heuristics for the MMCP with arbitrary graph structure.
\item {\textit{Graph Simplification.}} We clarify that target cut-edges can only be obtained on bridges or within connected components formed after breaking all bridges (Theorem~\ref{theorem:1}), which can simplify the scale of the graph and reduce the running time.
\item {\textit{Guaranteed Unsupervised.}} The cost and penalty of MMCP are explicitly written for the first time. An efficient unsupervised learning pipeline with a performance guarantee is designed to quickly find high-quality solutions that satisfy the constraint (Theorem \ref{theorem:3}).
\item {\textit{Solution Forest.}} We prove that all feasible solutions can be equivalently achieved by disconnecting an edge in one of the spanning trees of the graph, i.e., each solution corresponds to at least one spanning tree (Theorem \ref{theorem:4}).
\item {\textit{Novel Heuristics.}} A novel heuristic approach is proposed to repair and improve the solutions of the unsupervised solver that searches for better solutions by adding or removing vertices to realize a transformation of the spanning tree.
% \item {\textit{Remarkable Effect.}} We evaluate PIONEER in a wide variety of experiments. In particular, we manifest that the proposed method is able to reliably and efficiently find feasible solutions of good quality.
\end{itemize}

%The remaining part of the paper is structured as follows. Section 2 briefly introduces the related works. Section 3 is devoted to presenting the notation, the definition, and the inherent complexity of the maximum minimal cut problem. The details of the proposed PIONEER are described in Section 4. The experimental results and case studies are illustrated in Section 5. In Section 6, conclusions and possible directions for future work are given.

\section{Related Work}
This section briefly reviews the related work on the maximum minimal cut problem and the combinatorial optimization solvers.

\textbf{Maximum Minimal Cut Problem (MMCP).} Unlike general CO problems, the study of MMCP is still in its infancy. As a NP-hard issue, the optimal solution of MMCP cannot be approximated by a constant factor in polynomial time unless P = NP \cite{Duarte19}. Based on this point, researchers have developed a wide range of complexity analyses for the MMCP, while there exist rare results about the approximate solvers of the MMCP in general graphs \cite{Ding16, Aldred09}. Notably, Flynn et al. \cite{Flynn17} showed that the size of the largest bond is at least $\Omega(n^{\log_3{2}})$ for any simple 3-connected graph $G(V, E)$, where $n=|V|$.
%Aldred et al. \cite{Aldred09} showed that the maximum minimal cut of regular graphs whose degree are at least three is two vertex sets of the same size. 
Later, Duarte et al. \cite{Duarte21} proved that the MMCP is NP-complete even on planar bipartite graphs and split graphs. Moreover, they gave a $\mathcal{O}^*(2^{\mathcal{O}(tw\log{tw})})$-time algorithm for the MMCP from the perspective of the parameterized complexity. Although these methods have achieved breakthroughs in the computational theory of MMCP, they are still limited in the application of the real scene.
% Since the double connectivity requirements of cuts, the edges between $K$ and $L$ are at most $|E|-|V|+2$ for a graph $G=(V, E)$, where $(K, L)$ is the maximum minimal cut of $G$. In other words, a simple connected graph $G$ can be partitioned into two parts which can induce a tree, respectively.

\textbf{Combinatorial Optimization (CO) Solver.} Due to the high time complexity of exact algorithms \cite{pekny1990exact, rehfeldt2019combining}, approximation approaches have become the mainstream for solving CO problems \cite{gao2022dynamic, brodowsky2023approximation}. Heuristics are one of the most efficient and effective approaches for solving the CO problem which obtains a sub-optimal solution within a reasonable time \cite{Helsgaun00, Tang22, zhang2020seal}. Unfortunately, heuristics are problem-specific and time-consuming. With the blowout of artificial intelligence, the approaches of neural networks for CO emerge as the times require \cite{xin2021neurolkh, li2021learning, palm2018recurrent}. Most neural approaches to CO are supervised, and such methods rely on training large-scale labeled instances \cite{gasse2019exact, wang2019satnet}. In general, training unlabeled data is more challenging, which leads to two technical routes, i.e. reinforcement learning (RL) \cite{khalil2017learning, zheng2021combining} and unsupervised learning \cite{khalil2022mip, min2022can}. RL is known to be notoriously unstable to train. The works that are more relevant to ours are those unsupervised learning frameworks trained in an end-to-end manner. Wang et al. \cite{WangL23} adopt a relaxation-plus-rounding mechanism based on \cite{karalias2020erdos}, which guarantees performance by an entry-wise concave principle. However, the inability to write the loss of MMCP and the particularity of the connectivity prevents the application of the above methods.

\begin{figure*}[t]
\centering
\includegraphics[width=1\textwidth]{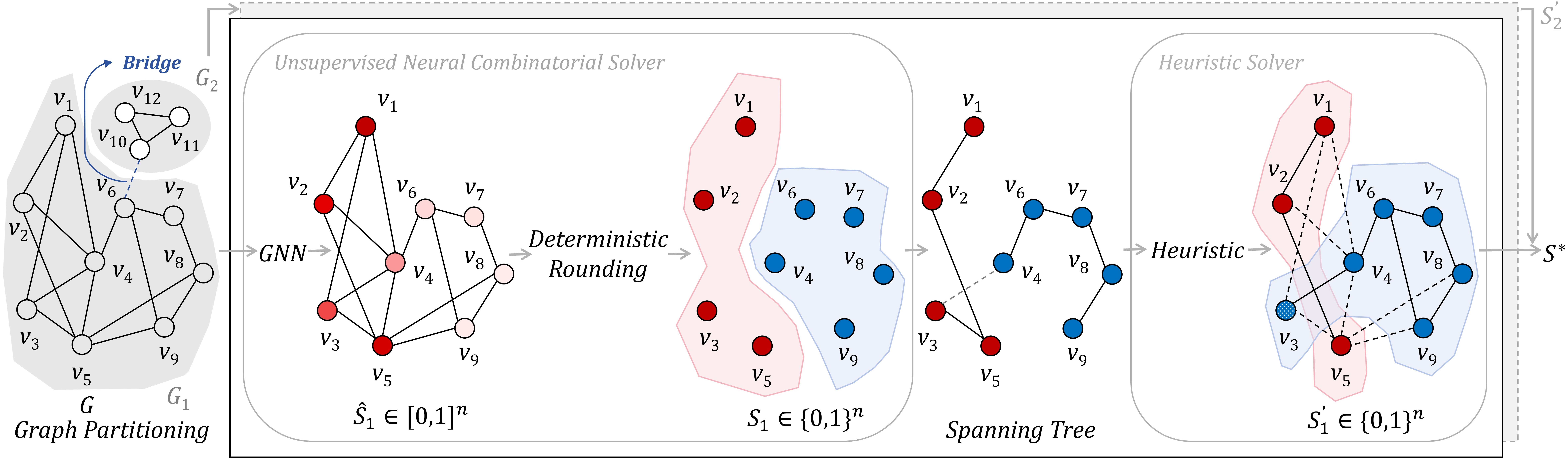}
\caption{Overview of the PIONEER pipeline. (a) Remove all bridges to induce connected subgraphs. (b) Utilize the unsupervised solver with a performance guarantee to obtain discrete solutions. (c) Construct spanning trees based on the discrete solutions. (d) Further enhance the solutions through the heuristic solver. (e) Compare all solutions to determine the optimal solution.} 
\label{fig:2}
\end{figure*}

\section{Preliminaries}
In this section, we describe the key concept applied in the paper. Firstly, We give some notations that will be used. 

\noindent\textbf{Notation.} Let $G(V, E, W)$ be a weighted, undirected, connected graph with $n=\left| V \right|$ vertices, $m=\left| E \right|$ edges, and weights $W=(\omega_1, …, \omega_m)$, where $\omega \in \mathbb{R^+}$ and $n,m \in \mathbb{N^*}$. Unweighted graphs can be regarded as weighted graphs with all edge weights equal to $1$. There are two disjoint connected subgraphs $K=(V_1, E_1)$ and $L=(V_2, E_2)$ of $G$, $K$ and $L$ can be connected by the edge set $F=(e_1, …, e_c)$ which has weight $W_F=(\omega'_1, …, \omega'_c)$. Besides, $\hat{S}$ and $S$ denote relaxed and discrete solutions throughout the paper.

Then, the problem that we are addressing, i.e. the maximum minimal cut problem (MMCP), is formulated.

\noindent\textbf{Problem Formulation.} The MMCP is to find a set of edges that divides the connected graph into two connected subgraphs so that the sum of the weight (or cardinality) on the edge set is the largest. The formal definition is given as follows.

\begin{definition}[The Maximum Minimal Cut Problem]
If $K$ and $L$ satisfy: 1) $V_1 \cup V_2 = V$ and $V_1 \cap V_2 = \emptyset$; 2) $E_1, E_2, F \subset E$, $E_1 \cup E_2 \cup F = E$ and $E_1 \cap E_2 \cap F = \emptyset$; 3) $c \in \mathbb{N^*}$.

Then $C=(K, L)$ can be called a minimal cut of $G$, and $F$ is named the cut-set of $G$. Alternatively, $K$ and $L$ are known as the connected cut. Each $C$ and $F$ corresponds uniquely, where the cut value of $\left| F \right|$ is given by $\sum\nolimits_{i=0}^c\omega'_i$. The maximum minimal cut is the minimal cut $C^{\ast}$ with the largest cut value among all $C$ of $G$.% Also worth noting is that $\left| F \right|=c$ for undirected graphs. 
\end{definition}

In the above definition, we can consider $F$ or $C_v $ as the solution of MMCP, where $C_v = (V_1, V_2)$ is the vertex set form of $C$. We assert that MMCP is more challenging than the Max-cut problem \cite{Haglin91}.

\section{Methodology}
In this section, the proposed framework, its components, and related theories will be elaborated comprehensively. Moreover, a random algorithm is presented as a baseline.%\footnote{roadmap of this section——limit to space limitations}

\subsection{Overview}
In summary, we achieve acceleration through graph partitioning and the performance-guaranteed unsupervised solver, and further repair and obtain higher-quality solutions by leveraging the heuristic solver. We overview the proposed unsupervised learning framework combined with heuristics (PIONEER) in Figure~\ref{fig:2}.  %\footnote{give a sketch flow here, suggest to split to components and steps——blow the figure1}

Given a connected graph $G = (V, E)$ as input, the connected components of $G$ can be formed by removing all bridges of $G$, which does not affect the final solution and also reduces the graph size (see Section \ref{subsection: partitioning}). The connected components with a larger number of vertices are solved jointly utilizing the proposed unsupervised solver (see Section \ref{subsection: unsupervised}) and heuristic solver (see Section \ref{subsection: heuristic}), while the solutions of smaller parts are obtained quickly by the exact algorithm. It is worth noting that all the solutions satisfying the constraints correspond to at least one spanning tree of $G$, which is the design source of the proposed heuristic solver (see Section \ref{subsection: solution}). %Our method fills a gap in effectively solving the maximum minimal cut problem (MMCP).

%\footnote{need connection}

Specifically, given $G$, we construct connected subgraphs $G_l$ and $G_s$ with a large and small number of vertices respectively according to graph partitioning (Theorem \ref{theorem:1}). The solution $S_s$ of $G_s$ is obtained by the brute force search algorithm whose cut value is $W_s$. 

Following the paradigm of unsupervised learning ~\cite{karalias2020erdos, wang2022unsupervised}, we adopts a relaxation-plus-rounding
approach. We optimize the loss function $\mathcal L$ to generate relaxed solutions $\hat{S}_l \in [0, 1]^n$, which is followed by a deterministic rounding to transform the solution in discrete space $\{0, 1\}^n$. The issue is whether the discrete solution $S_l$ can be achieved with assurance. Our key observation is that such success for MMCP essentially depends on how to explicitly write the loss and valid rounding. Therefore, one of our contributions is to expressly write the cost and constraint for MMCP. Furthermore, we show that the unsupervised solver can produce valid and low-cost solutions (Theorem \ref{theorem:3}) through reasonable rounding (Definition \ref{def:2}).

Despite the success of the proposed unsupervised solver on MMCP, an open question is how to repair the illegal solutions and further improve the quality of the solutions. Due to the fragile connectivity of sparse graphs and the strong correlation between connectivity and graph structure, relying only on network learning may not always result in solutions that satisfy the constraint. An important finding is that all feasible solutions correspond to at least one spanning tree of graph $G$ (Theorem \ref{theorem:4}). With this conclusion, we utilize the results of the unsupervised solver to construct spanning trees that are treated as the starting point of the heuristic solver for further exploration, and then we have new $S'_l$ and $W'_l$. Finally, the optimal solution $S^{\ast} = \arg\max_{S} \{W'_{l1}, ..., W'_{la}, W_{s1}, ..., W_{sb}\}$ with the largest cut value is selected as the final result, where $a$ and $b$ denote the maximum number of large and small subgraphs, separately.

In the rest of this section, we elaborately state the key theory and components of the proposed PIONEER, in order \emph{graph partitioning}, \emph{unsupervised combinatorial solver}, \emph{solution forest}, and \emph{heuristic tree transformation}. Then, we design a \emph{random tree search} as a baseline.

\subsection{Graph Partitioning}
\label{subsection: partitioning}
Considering that the time complexity of solving MMCP grows exponentially with the graph size, we exploit the theorem presented in this subsection to simplify the graph without affecting the solutions.

The bridge is a special edge in the undirected connected graph, also called cut-edge. The removal of the bridge will increase the number of connected components in the graph. Suppose that $G_1$ and $G_2$ are the connected components obtained by disconnecting one bridge $e_b$ of $G$, which satisfies $G_1 \cup G_2 \cup e_b = G$, $G_1 \cap G_2 = \emptyset$. In addition, $F, F_1$, and $F_2$ are the maximum minimal cut-set of $G, G_1$, and $G_2$, respectively. Then, we have the following theorem.
% \footnote{what is the role of the theorem——first paragraph}

\begin{theorem}
\label{theorem:1}
The maximum minimal cut-set $F$ of $G$ can only be obtained on one of the two sides of the bridge or on the bridge, that is, $F = max\{e_b, F_1, F_2\}$.
\end{theorem}

The above theorem manifests that the solution of MMCP for a connected graph $G$ can be obtained by separately solving connected subgraphs of $G$ which are divided by removing all bridges of $G$. We solve for connected subgraphs based on this graph partitioning instead of the whole graph. Therefore, the entire solving process can be accelerated due to the reduction of the graph scale.

\subsection{The Unsupervised Combinatorial Solver}
\label{subsection: unsupervised}
We aim to leverage unsupervised learning for solving MMCP. Generally, combinatorial optimization (CO) problems on graphs admit solutions that are binary vectors $S=\{x_1,...,x_n\} \in \{0,1\}^n$ of the set of vertices to denote whether the vertex is selected or not. Thus, a CO problem on graphs is to minimize a cost $f$ given a constraint $g$ by solving the following equation. 

\begin{equation}
\min\limits_{S \subseteq V} f(S;G) \enspace \text { subject to } \enspace g(S;G) \leq 1
\end{equation}

\begin{figure}[ht]
\centering
\includegraphics[width=1\columnwidth]{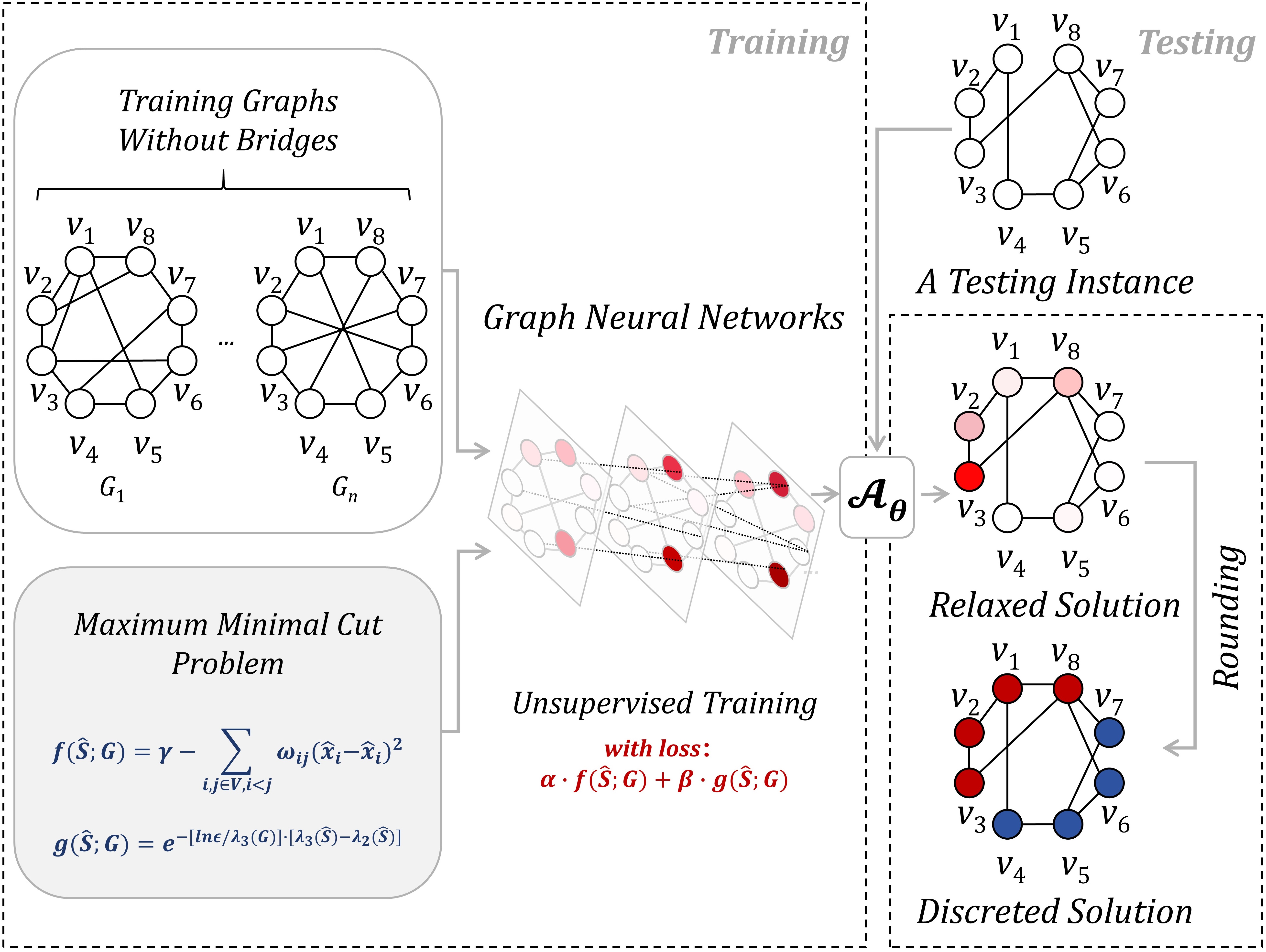}
\caption{The pipeline of unsupervised solver. The parameter $\theta$ is learned by graph neural networks, trained through graph instances without bridges. For testing, the network outputs relaxed solutions that are rounded to discrete solutions.}
\label{fig:3}
\end{figure}

\textbf{Learning for MMCP.} The learning for MMCP is to learn an algorithm $\mathcal{A}_\theta(G) \rightarrow S \in \{0,1\}^n$, e.g. a neural network (NN) parameterized by $\theta$ to solve MMCP. Given an undirected weighted graph $G(V, E,\omega)$ with $|V|=n$ and $|E|=m$, whose degree matrix and adjacency matrix are denoted by $D$ and $A$, respectively. The complete pipeline of unsupervised solver is given in Figure \ref{fig:3}. 

%\footnote{introduce the steps one by one?——below the Figure 3}}

Inspired by ~\cite{wang2022unsupervised}, we adopt a relaxation-plus-rounding mechanism and add a term to the loss function that penalizes deviations from the constraint. Let a continuous vector $\hat{S} = \{\hat{x}_1,...,\hat{x}_n\} \in [0,1]^n$ as the relaxed solution obtained by graph neural networks (GNNs). Formally, we define a new loss function for MMCP:

\begin{equation}
\min\limits_{\theta} \enspace \mathcal L(\theta; G) \triangleq  \alpha \cdot f(\hat{S};G) + \beta \cdot g(\hat{S};G)), \alpha, \beta > 0
\end{equation}

\noindent where, $\alpha = m/n$ or $1$, $\beta = \max f(S;G)$. Note that $\beta = \sum_{i=1}^{m} \omega_i$ for weighted graph and $\beta = m$ for unweighted graph. Since connectivity is a feature that is strongly correlated with the number of edges, $\alpha$ controls the balance between cost and penalty. %, which is beneficial to getting better solutions.

\underline{\emph{Cost Function.}} To ensure the non-negativity of the loss function, we translate the cost function by minimizing the difference between the maximum value $\gamma$ of the objective and the sum of cut-set weights. Typically, the sum of edge weights or cardinality of the graph is employed as $\gamma$. We define the cost function as follows. 

\begin{equation}
f(\hat{S};G) = \gamma - \sum_{i, j \in V, i < j} \omega_{ij} (\hat{x}_i - \hat{x}_j )^2
\end{equation}

\underline{\emph{Penalty Function.}} The connected constrain $g(\hat{S}; G))$ for MMCP is then considered. According to the definition of a maximal minimal cut, we devote to dividing the graph $G$ into two cuts by encoding all the vertices as $0$ or $1$. Thus, the constraint is satisfied if the graph $G'$, with the removal of edges between $0$ and $1$, has only two connected components; otherwise, it is illegal and should be punished. 

We enlighten from the Matrix-Tree theorem~\cite{ghosh2006growing} in the Appendix \ref{appendix:A5} and the property of Laplacian matrix, i.e. the number of eigenvalues with value $0$ of Laplacian matrix is equal to the number of connected components of $G$. Consequently, the number of eigenvalues with value $0$ of the Laplace matrix $L(G')$ for $G'$ should be equal to $2$. Formally, let $\lambda_1 \leq ... \leq \lambda_n$ is ordered eigenvalues of $L(G')$, there is $\lambda_1 = \lambda_2 = 0, \lambda_3 > 0$ if the solution satisfies the constraint. Note that $L=D-A$. Then, the adjacency matrix of $G'$ can be written by the relaxed solution $\hat{S}$: 

\begin{equation}
\label{eq:4}
A(G') = \left\{ 
    \begin{aligned}
    &0, & &a_{ii} \cr 
    &(1-\hat{x}_i-\hat{x}_j)^2 \cdot a_{ij}(G), & &a_{ij},i \neq j
    \end{aligned}
\right.
\end{equation}

\noindent where $a_{ij}(G)$ is the elements of $A(G)$, $a_{ii}$ and $a_{ij}$ are the elements of $A(G')$. The degree matrix can be built by summing the elements of each row in $A(G')$. Therefore, the corresponding eigenvalues of the relaxed solution can be easily obtained.

In addition, for graphs with different sizes, there is a significant difference in the magnitude of the eigenvalues of the Laplace matrix, which is detrimental to learning. Thus, we attempt to find an upper bound of the third small eigenvalue of $G'$ based on Theorem~\ref{theorem:2} which guarantees the robustness of the model. The proof of Theorem \ref{theorem:2} is shown in Appendix \ref{appendix:A2}.

\begin{theorem}[Eigenvalue Interlacing for Laplacian]
\label{theorem:2}
Let $L$ and $L'$ be Laplacian matrices of graph $G$ and its subgraph $G$ with respective ordered eigenvalues $\mu_1 \leq ... \leq \mu_n$ and $\nu_1 \leq ... \leq \nu_n$. Then the following inequality holds:
\begin{equation}
\nu_i \leq \mu_i, i=1,...,n.
\nonumber
\end{equation}
\end{theorem}

According to the above theorem, we can regard $\lambda_3(G)$ of $L(G)$ as an upper bound of $\lambda_3(G')$ of all $L(G')$ and denote $\lambda_3(G')$ as $\lambda_3(S)$ as well. Meanwhile, the exponential function is employed to ensure that the learning process is differentiable and the necessary monotonic relation. The following penalty function is defined: 

\begin{equation}
g(\hat{S};G)) = e^{-[ln\epsilon/\lambda_3(G)] \cdot [\lambda_3(\hat{S})-\lambda_2(\hat{S})]}
\end{equation}

where $\tau = ln\epsilon/\lambda_3(G) > 0$ is a self-adaptive coefficient for each graph, used to reconcile the differences in the upper bounds of the eigenvalues due to different sizes of graphs, and $\epsilon$ is a small value, e.g. $\epsilon = 10^{-4}$.

\textbf{Deterministic Rounding.} After network training, assuming that the optimized parameter $\theta$ enable $\mathcal L$ to be small, and we expect the relaxed solution $\hat{S}$ can be deterministically rounded to a valid discrete solution $S$ by sequential decoding as following forms.

\begin{definition}[Deterministic Rounding]
\label{def:2}
Given a continuous vector $\hat{S} = \{\hat{x}_1,...,\hat{x}_n\} \in [0,1]^n$, w.o.l.g., $i=1,...,n$. Set $\bar{S} = \arg\min_{j=0,1} \mathcal L(\bar{x}_1,...,j,...,\bar{x}_n)$, i.e. round $\hat{x}_i$ into $0$ or $1$ and fix all the other elements unchanged, then repeat the procedure until all the variables become discrete. If $\mathcal L(\hat{S}) \leq \mathcal L(\bar{S})$, then $\bar{x}_i$ is rounded to $1-\bar{x}_i$ to minimize the number of $0$ eigenvalues, which induces a new discrete solution $S$. Then, repeat the procedure until $\mathcal L(\hat{S}) > \mathcal L(S)$.
\end{definition}

\textbf{Performance Guarantee.} We prove the following theorem to show that our unsupervised framework allows generating a feasible and low-cost solution $S$ after deterministic rounding in theory. We move the proof to Appendix~\ref{appendix:A3} due to the interest of space.

\begin{theorem}[Performance Guarantee]
\label{theorem:3}
Let $\beta > \mathrm{max}_{S \in \Omega}f(S;G)$ and $\mathrm{min}_{S \in \Omega}f(S;G) \geq 0$. Suppose that the learned parameter $\theta$ achieves $\mathcal L(\theta,G) < \alpha \cdot \beta$. Then, rounding the relaxed solution $\hat{S} = \mathcal{A}_\theta(G)$ to a discrete solution $S \in \Omega$ such that $\alpha \cdot f(S;G) < \mathcal L(\theta;G)$.
\end{theorem}

The above theorem indicates that the loss will not increase after the deterministic rounding. On this basis, once the parameter $\theta$ gets optimized to $\mathcal L(\theta; G)<\alpha \cdot \beta$, there is $\alpha \cdot f(S;G) + \beta \cdot g(S;G) \leq \mathcal L(\theta; G) < \alpha \cdot \beta$. Owing to $f(\cdot) \geq 0$ and $g(\cdot) \geq 0$, we have $\alpha \cdot f(S;G)<\mathcal L(\theta; G)$, s.t $g(S;G) \leq 1$. 

\textbf{Constraint-prior Rounding.} Furthermore, as that connectivity might pose more challenges for network learning, to address potential infeasible solutions and prioritize the assurance of solution feasibility, we employ an additional constraint-prior rounding, i.e. set $S = \arg\max_{j=0,1} \lambda_3(x_1,...,j,...,x_n)$ for the illegal solutions after the aforementioned deterministic rounding.

% \footnote{how about connect to the next section?}

\subsection{Solution Forest}
\label{subsection: solution}

A natural idea for addressing CO problems is to initially discover a solution and subsequently verify whether the constraints are met. However, if the solution adhering to the constraints can be reformulated into a directly obtainable equivalent form, it would substantially enhance the efficiency of the search. Consequently, we attempt to convert the Bernoulli solution of the MMCP into another closed form such that all solutions achieved by a given algorithm must satisfy the constraints. 

It is well-known that the spanning tree of the graph possesses some interesting properties \cite{zhang2006monkey}, such as, each edge of the tree is a bridge. Based on this conclusion, we state the following theorem. 

\begin{theorem}
\label{theorem:4}
Suppose that $C=(K, L)$ is a maximum minimal cut with cut value $\psi$ of $G$, which can definitely be obtained by disconnecting an edge of a certain spanning tree $T=(V_T, E_T)$ of $G$.
\end{theorem}

The theorem above indicates that each feasible solution of MMCP corresponds to at least one spanning tree of $G$. In other words, we can explore legal solutions for the MMCP by transforming the spanning tree, eliminating the need to check for constraint violations, which is the core idea behind the design of our subsequent heuristic solver. Additionally, to facilitate the subsequent description of the heuristic solver, we define a notion of disconnect-vertex as follows. 

\begin{definition}[Disconnected-Vertex]
An optimal solution of $G$ can be obtained by disconnecting the edge $e_d=(v_i,v_j)$ of spanning tree $T$. Such edge is called disconnected-edge of $T$ and $v_i$, $v_j$ are named disconnected-vertex of $T$.
\end{definition}

\subsection{Heuristic Tree Transformation}
\label{subsection: heuristic}

% \footnote{suggest to tell the goal, the challenge and basic idea——OK!}
To further repair and improve the solutions of the unsupervised solver while disregarding the impact of the graph structure, we aim to design a heuristic that achieves the transformation of the spanning tree by adding vertices for the cut to explore better solutions.

Suppose $C_v = (V_1, V_2)$ is the maximum minimal cut of $G$ obtained from the spanning tree $T$ of $G$, where $V_1 = V \backslash V_2$, w.l.o.g, we assume $\left| V_1 \right| \textless \left| V_2 \right|$. Let $p_k \in V_1$ and $p_{k+1} \in V_2$. We aim to optimize the existing spanning tree by introducing an appropriate $p_{k+1}$ related to $p_k$ into $V_1$, seeking enhanced solutions. The transformation of the current spanning tree will undergo the following four phases.

\textbf{\emph{Phase \uppercase\expandafter{\romannumeral1}. Selection}}.

As not all vertices in $V_2$ are suitable to join $V_1$, the selection of $p_{k+1}$ must meet the constraint that $p_{k+1}$ is the neighbor of $p_k$, which ensures $V_1$ is connectable after each vertex movement. Therefore, different neighbors of $V_1$ can be chosen to join $V_1$, altering the shape of the spanning tree in conjunction with the subsequent phases.

\textbf{\emph{Phase \uppercase\expandafter{\romannumeral2}. Disconnect edge}}.

Generally speaking, the selected vertex $p_{k+1}$ may also have other neighbors. Therefore, we disconnect the edges connected to $p_{k+1}$ in $T$ to isolate $p_{k+1}$, making it convenient for the subsequent addition of edges. The specific cases that arise when the edges related to $p_{k+1}$ are broken are discussed in Appendix~\ref{B:PIONEER}.

\textbf{\emph{Phase \uppercase\expandafter{\romannumeral3}. Add edge}}.

After disconnecting all feasible edges of $p_{k+1}$, the current graph consists of several connected components centered around $p_{k+1}$. At present, the edge addition involves those of $p_{k+1}$ and its neighbor $n$. For the addition of the edge connected to $p_{k+1}$, we directly link $(p_k, p_{k+1})$ to incorporate $p_{k+1}$ into $V_1$. Note that the current connected component containing $V_1 \cup P_{k+1}$ forms a tree, referred to as $T_b$. Handling the edge addition for $n$ introduces additional complexity and requires the following processing.

\underline{\emph{Candidate.}} The edges that need to be added are not arbitrary, they have to meet certain constraints. (i) The connected edges belong to $G$. (ii) The vertices planned to be connected must belong to $V_2$. (iii) Only one edge is added to the neighbor of each $P_{k+1}$ in $V_2$. (iv) There is no self-loop. The vertices that satisfy the above conditions are selected as the candidate set when adding edges.

\underline{\emph{Addition.}} After selecting the vertex, we connect it with $p_{k+1}$. However, an unfortunate observation arises that the coupling relationship between different connected components results in high time complexity when adding edges. The main objective of adding edges is to recombine discrete connected components to induce a new spanning tree. To address this issue, we construct the spanning tree $T_{be}$ using $V_2 \backslash p_{k+1}$, providing a new way to achieve the same effect of adding edges in practice. Notably, $V_2 \backslash p_{k+1}$ can be capable of constructing a spanning tree as there is no bridge after graph partitioning.

\underline{\emph{Montage.}} It is evident that the current graph consists of two trees, containing $V_2 \backslash p_{k+1}$ and $V_1 \cup p_{k+1}$, respectively. We concatenate $T_b$ and $T_{be}$ while preserving the original tree edge of $T$ connecting $V_1$ and $V_2$, ensuring that the graph obtained after transformation remains a spanning tree $T_{new}$ of $G$.

\textbf{\emph{Phase \uppercase\expandafter{\romannumeral4}. Dislodge vertex}}. 

Note that, for $(V_1 \cup p_{k+1}, V_2 \backslash p_{k+1})$, it is not necessarily the optimal solution for $T_{new}$. The optimal solution $C^{\ast}_v = (V^{\ast}_1, V^{\ast}_2)$ of $T_{new}$ may potentially be found at other disconnected-edges. Additionally, an observation is that removing some vertices from $V^{\ast}_1$ might lead to a greater increase in cut value. Hence, we move vertices of $V^{\ast}_1$ to $V^{\ast}_2$ in turn to explore better solutions. The movement is restricted to the vertices in $V^{\ast}_1$ who have neighbors in $V^{\ast}_2$.

It is noteworthy that the idea of transforming spanning trees to search for better solutions can be a general scheme for solving MMCP, which is worth further exploration in the future.

\subsection{Random Tree Search}
\label{subsection:rts}
According to Theorem~\ref{theorem:4}, we designed a random search algorithm as a baseline based on the conclusion that transforming the shape of the spanning tree can lead to diverse feasible solutions. Drawing inspiration from the Kruskal algorithm \cite{kleinberg2006algorithm} employed in constructing minimum spanning trees, we arrange all edges in ascending order based on their weights. For unweighted graphs, the order of edges is randomized. In each iteration, the edge sequence is randomly shuffled, and subsequently, the spanning tree $T_r$ of the graph $G$ is constructed based on the shuffled edge sequence. Following this, a feasible solution for $G$ is obtained by disconnecting a single edge from $T_r$. The feasible solution $S_{T_r}$ with the maximum cut value is then selected as the optimal solution $S^{\ast}$ for $T_r$. The demonstration and the pseudo-code of the algorithm are given in Appendix \ref{appendix:B4}.

\section{Experimental Evaluation}
This section discusses our experimental setup and results. As a case study, we provide a specific real-world application of MMCP.

\subsection{Experimental Setup}
\label{subsection:setup}
We conduct extensive experiments to evaluate the practical effectiveness of the proposed method. Here we briefly describe the hardware, datasets, baselines, and evaluation metrics. 

\noindent\textbf{Implementations.} We implement the PIONEER by Python 3.7.11 and PyTorch 1.9.2. All experiments are conducted on an Ubuntu machine with NVIDIA GeForce RTX 3090 (24 GB VRAM) and Intel(R) Xeon(R) Platinum 8260 CPU @ 2.40GHz (256 GB RAM). For the unsupervised solver, the training, validating, and testing datasets use an 8:1:1 dataset split, and each synthetic dataset contains $10,000$ graphs. In addition, in the interest of fairness, our unsupervised solver and \cite{karalias2020erdos} share the same neural network structure and parameter setting without the GAT layer. We refer interested readers to \cite{karalias2020erdos} for more details. Given the significant variations and sparsity observed in real-world datasets, we select the suitable $\alpha$  based on parameter study in Appendix~\ref{subsection:sa}. We set $\alpha = m/n$ for synthetic graphs, allowing the model to explore the solutions that yield larger objective values. For the random tree search, we report the best result by running the algorithm for $500$ rounds. Our code can be available at \url{https://github.com/luckyseasalt/PIONEER}.

\noindent\textbf{Datasets.} We assess our methods in a wide variety of experiments, rigorously conducted on synthetic and real-world datasets. Real-world datasets are all unweighted graphs, including, ENZYMES \cite{you2021identity}, IMDB \cite{KKMMN2016}, and REDDIT \cite{yanardag2015deep} datasets. Noteworthy, the real-world dataset also includes actual power grids and associated parameters. 

\begin{figure}[h]
\centering
\includegraphics[width=1\columnwidth]{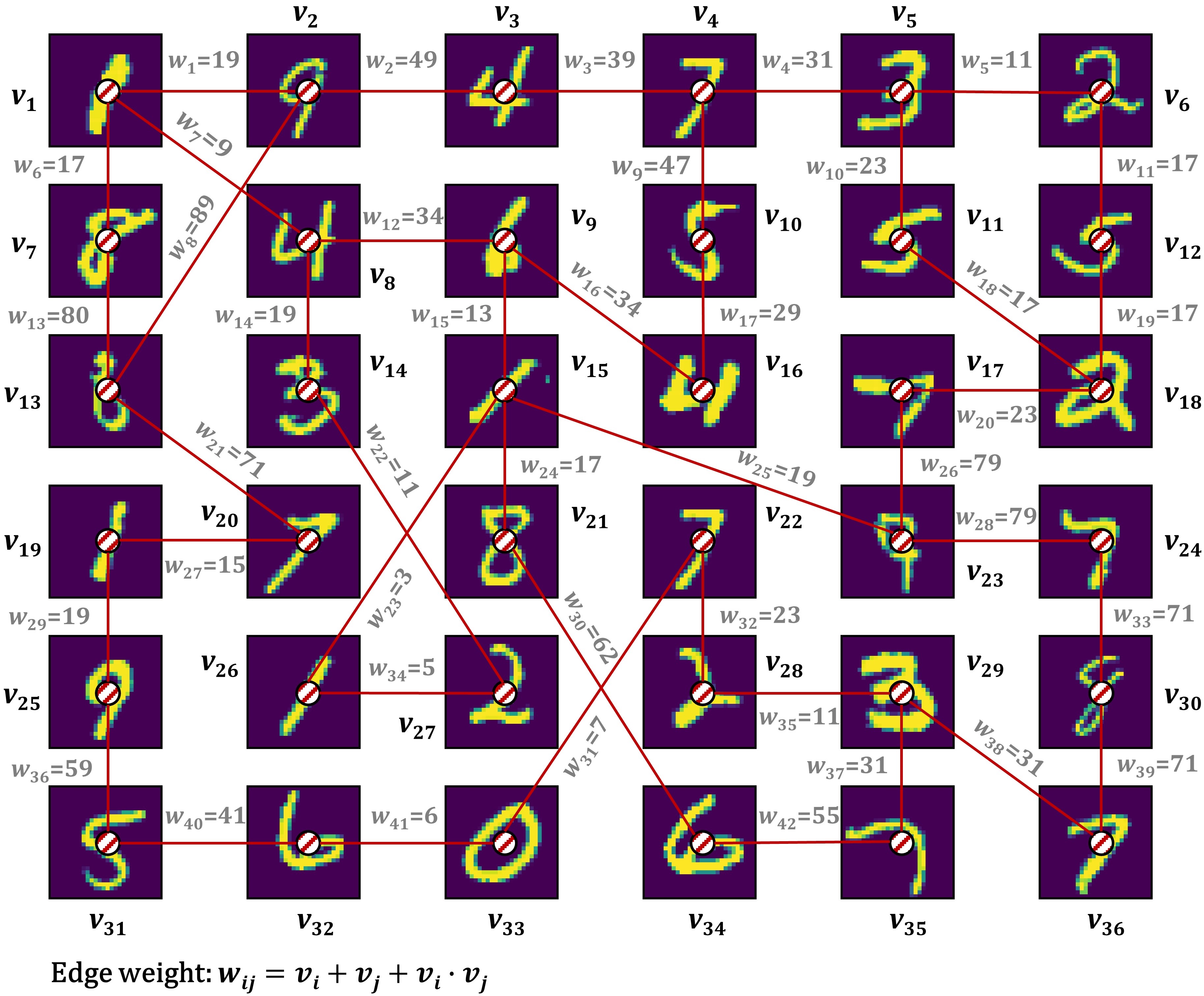}
\caption{An instance to show the synthetic datasets. The vertices in the graph correspond to random pictures of Mnist, whose values $v_i$ are the one-digit numbers that match the pictures. And the weight of edge $e_{ij}$ is equal to $v_i + v_j +v_i \cdot v_j$.}
\label{fig:4}
\end{figure}

Synthetic datasets are constructed with $36$ vertices based on the pictures in MNIST \cite{deng2012mnist}. There are $60,000$ pictures in Mnist which correspond to different one-digit \cite{deng2012mnist}. The construction of the synthetic datasets is inspired by \cite{poganvcic2019differentiation}. Here, we first generate a complete graph $\mathcal{G}$ as a motherboard. We associated each edge with $\phi = \{0, 1\}^{|E|}$ and each vertex with the picture. Thus, synthetic graphs can be built by a random sequence $\phi$ whose element denotes whether the corresponding edge in $\mathcal{G}$ exists. We give an instance in Figure \ref{fig:4}.

Considering the impact of various graph structures and the requirements of our framework, constraints can be imposed on the random graph generation process to obtain different types of graphs. It should be noted that the unsupervised solver is employed to handle graphs that have undergone graph partitioning. Hence, we focus on the scenario without bridges when creating the dataset. In summary, the synthetic Mnist datasets are classified into two primary types, i.e. isomorphic graphs and heterogeneous graphs with the notation `I-36|$m$' and `H-36', where $m$ is the number of edges. More configurations of all datasets are described in Appendix~\ref{appendix:C1}.

\noindent\textbf{Baselines.}
As few algorithms can be fully adapted to solve MMCP, we mainly designed the brute force algorithm and the random algorithm, which can individually obtain the exact and approximate solutions. In addition, we adaptively modified the genetic algorithm \cite{benabbou2020interactive}. We leave more details of these baselines in Appendix \ref{B:details}.

\noindent\textbf{Metrics.} Consistent with existing studies, all results are measured by widely used metrics. Unfortunately, due to the unavailability of labeled datasets that offer accurate solutions to MMCP, we resort to the value of cut-set and execution time to compare the different approaches. Assume that the solution is $S \in \{0,1\}^n$, and the edges between the labeled vertices $0$ and $1$ is $F$ with weight $\omega_1,...,\omega_m$, then the value of the cut-set is $W_F = \sum_{i=1}^m \omega_i$. Execution time is measured in sec. per graph (s/g). We record the mean and standard deviation by running the results for $5$ times.

\subsection{Main Results}
\label{subsection:mr}

Table~\ref{tab:smr}, \ref{tab:rwd} and \ref{tab:rwr} report the results on synthetic datasets and real-world datasets, respectively. Note that the results of the brute force algorithm are not included because it is still too time-consuming even when adopting graph partitioning to simplify. All methods are graphically simplified and accelerated by graph partitioning.

In summary, the proposed PIONEER achieves better solutions than the competitors on all datasets and runs quickly. The results show that PIONEER exhibits superior performance in solving MMCP with various graph structures. We discuss the results of each task in more detail below. Note that the best results among the methods are highlighted in bold across all Tables.

\noindent\textbf{Experiments on synthetic datasets.} The results of the synthetic tasks are shown in Table~\ref{tab:smr}, where PIONEER demonstrates competitive performance across all datasets. It consistently delivers solutions of the highest quality while maintaining a rapid processing speed. Surprisingly, the random tree search method achieves acceptable results within a relatively short time frame, benefiting from Theorem~\ref{theorem:4}. However, the performance of the proposed random algorithm weakens as the number of edges increases. This phenomenon is attributed to the significant increase in the number of spanning trees for the graphs with more edges, limiting its ability to find higher-quality solutions. In contrast, PIONEER does not rely on edge sequences for spanning tree construction. Instead, it achieves tree transformation by adding vertices, effectively overcoming this limitation and finding better solutions.

\begin{table}[h]
  \caption{Performance comparison on synthetic datasets.}
  \label{tab:smr}
  \centering
  \begin{tabular}{ccc}
    \hline
    Dataset & Random Tree & \textbf{PIONEER} \\
    \hline
    I-36|60 & \makecell{726.84 $\pm$ 0.49 (2.74 s/g)}& \makecell{\textbf{872.34 $\pm$ 0.76} (0.95 s/g)} \\
    I-36|120 & \makecell{1864.13 $\pm$ 2.22 (2.87 s/g)}& \makecell{\textbf{2374.61 $\pm$ 2.43} (3.19 s/g)} \\
    I-36|180 & \makecell{2886.29 $\pm$ 3.74 (2.99 s/g)}& \makecell{\textbf{3543.06 $\pm$ 3.22} (5.76 s/g)} \\
    I-36|240 & \makecell{3881.32 $\pm$ 2.69 (3.09 s/g)}& \makecell{\textbf{4573.49 $\pm$ 4.17} (6.46 s/g)} \\
    I-36|300 & \makecell{4861.74 $\pm$ 2.40 (3.22 s/g)}& \makecell{\textbf{5514.47 $\pm$ 4.79} (8.45 s/g)} \\
    Heter-36 & \makecell{5033.12 $\pm$ 2.44 (3.27 s/g)} & \makecell{\textbf{5659.22 $\pm$ 6.43} (8.46 s/g)} \\
    \hline
\end{tabular}
\end{table}

\noindent\textbf{Experiments on real-world datasets.} To further demonstrate the superiority of our PIONEER, we perform experiments on real-world datasets. We note that the graphs under consideration exhibit sparsity and the presence of bridges, rendering graph partitioning highly efficacious. Table~\ref{tab:rwd} reports the statistics of the real-world datasets and the results of graph partitioning. More details on the processing of the dataset are in the Appendix \ref{appendix:C2}. We categorize graphs containing fewer than $16$ vertices as small graphs and employ a brute force algorithm to precisely determine their values. Otherwise, we employ the PIONEER method to address MMCP.

\begin{table}[H]
  \caption{Statistics of the real-world datasets. Since each dataset consists of many graphs, we counted the range of values of the corresponding variables.}
  \label{tab:rwd}
  \centering
  \resizebox{1\columnwidth}{!}{
  \begin{tabular}{ccccccc}
  \hline
  \multirow{2}{*}{Dataset} & \multirow{2}{*}{Type} & \multirow{2}{*}{\# Vertex} & \multirow{2}{*}{\# Edge} & \multicolumn{2}{c}{\textbf{\# Subgraph }}\\
  ~ & ~ & ~ & ~ & $n \leq 16$ & $n > 16$\\
  \hline
  \multirow{4}{*}{ENZYMES} & Origin & [2, 126] & [1, 149] & \multicolumn{2}{c}{575} & ~ \\
  ~ & Train & [16, 116] & [27, 144] & 385 & 337 \\
  ~ & Valid & [18, 66] & [36, 120] & 52 & 46 \\
  ~ & Test & [16, 60] & [31, 114] & 75 & 39 \\
  \multirow{4}{*}{IMDB} & Origin & [12, 136] & [52, 2498] & \multicolumn{2}{c}{1000} & ~ \\
  ~ & Train & [16, 136] & [35, 1249] & 358 & 442 \\
  ~ & Valid & [16, 49] & [36, 445] & 37 & 63 \\
  ~ & Test & [16, 55] & [39, 785] & 51 & 49 \\
  \multirow{4}{*}{REDDIT} & Origin & [6, 3782] & [8, 8142] & \multicolumn{2}{c}{2000} & ~ \\
  ~ & Train & [16, 711] & [21, 1631] & 593 & 1493 \\
  ~ & Valid & [16, 86] & [20, 165] & 123 & 75 \\
  ~ & Test & [16, 112] & [21, 194] & 127 & 85 \\
  \hline
\end{tabular}}
\end{table}

In Table~\ref{tab:rwr}, we can see that PIONEER outperforms the baselines across all datasets. Specifically, there is no statistically significant difference in terms of their performance on the IMDB and REDDIT, while PIONEER requires less time. We would like to emphasize that the random algorithm works wonders for certain tree-structured graphs, whereas PIONEER offers a versatile framework that can be applied to various graph structures. It should be noted that PIONEER achieves significantly larger solutions in terms of addressing the MMCP of ENZYMES, which reveals that even in sparse tree-structured graphs, our method can still find superior solutions effortlessly.

\begin{table}[h]
  \caption{Performance comparison on real-world datasets.}
  \label{tab:rwr}
  \centering
  \begin{tabular}{ccc}
    \hline
    Dataset & Random Tree & \textbf{PIONEER} \\
    \hline
    ENZYMES & \makecell{19.85 $\pm$ 0.32 (3.49 s/g)}& \makecell{\textbf{31.32 $\pm$ 0.43} \textbf{(2.46 s/g)}} \\
    IMDB & \makecell{56.65 $\pm$ 0.03 (1.63 s/g)} & \makecell{\textbf{56.70 $\pm$ 0.10 (1.08 s/g)}}\\
    REDDIT & \makecell{16.42 $\pm$ 0.05 (3.27 s/g)} & \makecell{\textbf{17.97 $\pm$ 0.02} \textbf{(1.58s/g)}}\\
    \hline
\end{tabular}
\end{table}

\subsection{Ablation Study}
To validate the effectiveness of the key components in PIONEER, we conduct ablation studies on all aforementioned datasets. Only two critical results are reported here.

\begin{figure}[h]
\centering
\includegraphics[width=0.9\columnwidth]{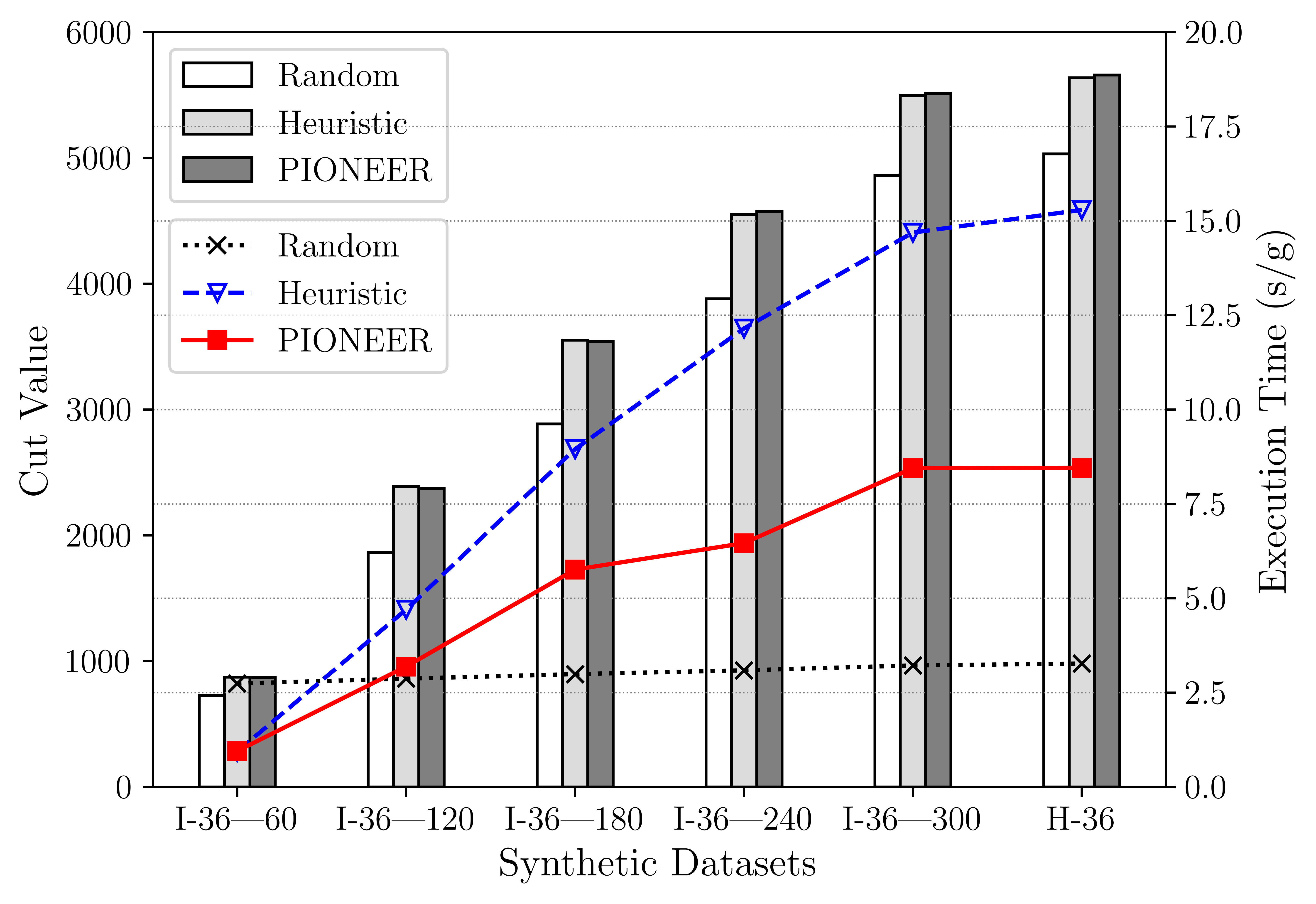}
\caption{Effectiveness of the unsupervised solver. We report the results on the synthetic dataset here. The bars denote the cut values, and the lines denote the execution time.}
\label{fig:5}
\end{figure}

\noindent \textbf{w/o Unsupervised solver.} We remove the unsupervised solver in our framework and directly construct a random spanning tree as input for the heuristic solver to evaluate the effectiveness of the unsupervised solver. As can be seen in Figure~\ref{fig:5}, the variant without an unsupervised solver can solve with almost the same quality as PIONEER (Avg. 3750.48 vs 3756.20). However, the speed of implementation of the two showed marked differences (Avg. 9.45 s/g vs 5.54 s/g). This is an interesting finding that the unsupervised solver serves as an effective starting point for the heuristic solver, leading to a substantial improvement in search speed.

\noindent \textbf{w/o Heuristic solver.} Similarly, we assess the heuristic by directly taking the output of the unsupervised solver as a result. The results are summarized in Table~\ref{tab:ur}, showing the heuristic solver plays a crucial role in repairing and enhancing the solutions. Encouragingly, although the unsupervised solver does not yield optimal solutions, its fastest running speed and acceptable solution quality are also impressive. Furthermore, we have observed that unsupervised solvers exhibit surprisingly effective performance in tackling MMCP for dense graphs with fast speed. 

\begin{table}[h]
  \caption{Effectiveness of the heuristic solver. The cut values are reported by mean values for correctly solved instances.}
  \label{tab:ur}
  \centering
  \begin{tabular}{cccc}
    \hline
    \multirow{2}{*}{Dataset} & \multicolumn{2}{c}{Unsupervised Solver } & \multirow{2}{*}{\textbf{PIONEER}}\\
    ~ & Violation & Cut Value & ~ \\
    % \textbf{Datasets}&  \makecell{Constraint\\Violation} & \makecell{Unsupervised\\Solver} & \textbf{PIONEER}\\
    \hline
    I-36|60 & 4.52 \% & \makecell{551.49 \textbf{(0.13 s/g)}} & \makecell{\textbf{872.34} (0.95 s/g)} \\
    I-36|120 & 0 \% & \makecell{2069.82 \textbf{(0.13 s/g)}} & \makecell{\textbf{2374.61} (3.19 s/g)} \\
    I-36|180 & 0 \% & \makecell{3363.73 \textbf{(0.41 s/g)}} & \makecell{\textbf{3543.06} (5.76 s/g)} \\
    I-36|240 & 0 \% & \makecell{4319.44 \textbf{(0.14 s/g)}} & \makecell{\textbf{4573.49} (6.46 s/g)} \\
    I-36|300 & 0 \% & \makecell{5222.96 \textbf{(0.14 s/g)}} & \makecell{\textbf{5514.47} (8.45 s/g)} \\
    H-36 & 0 \% & \makecell{5372.27 \textbf{(0.14 s/g)}} & \makecell{\textbf{5659.22} (8.46 s/g)} \\
    \hline
    ENZYMES & 0 \% & \makecell{12.62 \textbf{(0.17 s/g)}} & \makecell{\textbf{31.32} (2.46 s/g)} \\
    IMDB & 4.90 \% & \makecell{41.99 \textbf{(0.07 s/g)}} & \makecell{\textbf{56.70} (1.08 s/g)}\\
    REDDIT & 1.65 \% & \makecell{8.07 \textbf{(0.16 s/g)}} & \makecell{\textbf{17.97} (1.58 s/g)} \\
    \hline
\end{tabular}
\end{table}

\subsection{More Results} 
We conduct more experiments to further analyze the key components and parameters of PIONEER. For detailed information, please refer to Appendix \ref{appendix:C2} due to space limitations.

\textbf{Supplemental ablation study.} To validate the effectiveness of graph partitioning, rounding, and the upper bound $\lambda_3(G)$, additional ablation experiments are performed. The results indicate that graph partition can dramatically accelerate execution, the proposed rounding methods improve the ability to generate constraint-satisfying solutions, and $\lambda_3(G)$ enables the model to better learn the bi-connectivity for different graph structures.

\textbf{Parameter study.} Parameter experiments are conducted to study the key parameters, including the balanced control parameter $\alpha$ and the self-adaptive coefficient $\epsilon$. The results show that suitable $\alpha$ and $\epsilon$ can enhance the learning process of the unsupervised solver, leading to improved quality and legality of the solutions. 

\textbf{Performance guarantee study.} Our method avoids the intricate analysis of loss monotonicity, ensuring Theorem~\ref{theorem:3} through deterministic rounding. When the relaxed loss is successfully minimized to a sufficiently small value, the inequality $\mathcal L(S; G) < \mathcal L(\hat{S}; G)$ $< \alpha \cdot \beta$ is satisfied, yielding low-cost and feasible solutions.

\textbf{Heuristic study.} A constraint checking is incorporated into a genetic algorithm to perform a comparative study. The algorithm is provided in Appendix \ref{appendix:B5}. The results reveal that exploring solutions from spanning trees is significantly superior to the method of obtaining cuts and subsequently determining their legality.

%From the above comprehensive analysis, we can conclude that all the components involved are essential and effective.

\subsection{Case Study}
\label{subsec:case}
Recall that we mentioned a demand for the power grid motivating this work. Therefore, we utilize the task of cross-section identification in the power grid as a case study. We leave the problem statement and the visualization of solutions in Appendix~\ref{subsection:cs}. 

We study the performance of PIONEER on the power grid of certain regions. Due to the confidentiality of power grid data, we only give some important details in the Appendix \ref{appendix:C1}. Additionally, training is not feasible due to the limited dataset containing only three graphs. Therefore, we utilize models trained on the REDDIT dataset, which exhibits relative similarity to the power grid.

\begin{table*}
  \caption{Statistics and performance comparison on the task of the power grid. The numbers in ($\cdot$) are highlighted in underlined to indicate the maximum number of vertices in the subgraphs after graph partitioning.}
  \label{tab:pg}
  \centering
  \begin{tabular}{ccccccccc}
  \hline
  \multirow{2}{*}{\makecell{Code of \\ power grid}} & \multirow{2}{*}{\# Vertex} & \multirow{2}{*}{\# Edge} & \multirow{2}{*}{\# Bridge} & \multicolumn{2}{c}{\# Subgraph } & \multirow{2}{*}{Brute Force} & \multirow{2}{*}{Random Tree} & \multirow{2}{*}{PIONEER}\\
  ~ & ~ & ~ & ~ & $n \leq 16$ & $n > 16$ & ~ & ~ & ~\\
  \hline
  36-vertices & 36 & 42 & 12 & 0 & 1 (\underline{24}) & \textbf{26.12} (328.43 s/g) & \textbf{26.12} (1.14 s/g) & \textbf{26.12} (1.57 s/g)\\
  IEEE118 & 118 & 186 & 9 & 0 & 1 (\underline{109}) & / & 1831.67 (63.21 s/g) & \textbf{2659.34} (13.37 s/g)\\
  IEEE300 & 300 & 744 & 90 & 2 & 1 (\underline{206}) & / & 3610.43 (223.83 s/g) & \textbf{4151.21} (56.48 s/g)\\
  Area 1 & 2582 & 3028 & 1466 & 1 & 1 (\underline{1079}) & / & 169.55 (2516.48 s/g) & \textbf{276.92} (8604.68 s/g)\\
  Area 2 & 2799 & 3161 & 1794 & 36 & 5 (\underline{328}) & / & 50.83 (1635.30 s/g) & \textbf{59.28} (142.56 s/g)\\
  \hline
\end{tabular}
\end{table*}

As shown in Table \ref{tab:pg}, our method exhibits superior performance across all cases. Notably, PIONEER appears to take more time on Area 1 primarily due to the insufficient training of the unsupervised solver, resulting in the heuristic requiring more time for exploration. In contrast, relying solely on the heuristic solver requires a time of $11141.55$ s/g, indicating that even in this extreme case, the unsupervised solver can still contribute to a speedup for the heuristic.

\section{Conclusions and Future Works}

In this paper, we propose a novel unsupervised learning framework combined with heuristics named PIONEER for MMCP. In particular, we demonstrate the theoretical foundation of graph simplification and the equivalent form of solutions for MMCP. On these bases, we design an unsupervised framework with mathematical guarantees, which is not only sufficiently rapid in acquiring acceptable solutions independently but also enhances the speed of downstream solvers. We also construct a heuristic solver for further repairing and improving the quality of solutions. Extensive experiments manifest that PIONEER outperforms the baselines in various graph structures with good generalization. In the future, one promising direction is to enhance the ability of the unsupervised solver to handle sparse graphs. Another interesting theme is to explore a more explicit relationship between spanning trees and optimal solutions.

%\clearpage

%%
%% The acknowledgments section is defined using the "acks" environment
%% (and NOT an unnumbered section). This ensures the proper
%% identification of the section in the article metadata, and the
%% consistent spelling of the heading.

\begin{acks}
This paper was supported by the Science and Technology Project of State Grid: Research on artificial intelligence analysis technology of available transmission capacity (ATC) of the key section under multiple power grid operation modes (5100-202255020A-1-1-ZN).
\end{acks}

%%
%% The next two lines define the bibliography style to be used, and
%% the bibliography file.
\bibliographystyle{ACM-Reference-Format}
\bibliography{kdd2024}

%%
%% If your work has an appendix, this is the place to put it.
\clearpage

\appendix

\section*{Appendix}

In the Appendix, we provide proofs of the proposed theorems, details and pseudo-cod of the algorithms involved, more details of the experiments and more results, broader impact for the maximum minimal cut problem (MMCP), and license for all datasets. 

\section{The Proof and Related Theorem}
\label{appendix:A}
\subsection{Proof of Theorem \ref{theorem:1}}
\begin{proof}
Consider whether $e_d$ is in the cut-set $F$ of $G$. If $e_d \in F$, such that $F = e_d$, that is the maximum minimal cut $C = (G_1, G_2)$. Without loss of generality, we let $C = (K, L)$ and $e_d \in K$ if $e_d \notin F$. Suppose $v_1, v_2$ are on the different sides of $e_d$, and $v_1, v_2$ cannot be connected if $v_1,v_2 \in L$. Thus, the cut-set $F$ of $G$ will be equal to the $F$  of $G_1$ if $v_1,v_2 \in K$. Similarly, the case where $e_d \in L$ can also be proved. 
\end{proof}

\subsection{Proof of Theorem \ref{theorem:2}}
\label{appendix:A2}
\ 
\newline
To prove Theorem \ref{theorem:2}, we first introduce the eigenvalue interlacing theorem ~\cite{horn2012matrix}. 

\begin{theorem}[Weyl's inequality]
\label{theorem:5} Let $H$ and $U$ be $n \times n$ Hermitian matrices, with their ordered eigenvalues $\lambda_1(\cdot) \leq ... \leq \lambda_n(\cdot)$. Then the following inequality holds:
\begin{equation}
\lambda_i(H) + \lambda_n(U) \leq \lambda_i(H + U) \leq \lambda_i(H) + \lambda_1(U), i = 1,...,n.
\nonumber
\end{equation}
\end{theorem}

Endow $\mathbb{C}^n$ with the Euclidean inner product $\langle \cdot,\cdot \rangle$ which is anti-linear in the second coordinate. For $u \in \mathbb{C}^n$, the linear rank one map $u \otimes u = uu^\ast$. 

Then, we can draw the following corollary. 

\begin{corollary}
\label{corollary:1}
Let $H$ be a $n \times n$ Hermitian matrices with ordered eigenvalues $\lambda_1 \leq ... \leq \lambda_n$. For nonzero vector $u \in \mathbb{C}^n$ and $n \geq 2$, let the eigenvalues of $Z=H+u \otimes u$ be $\mu_1 \leq ..\leq \mu_n$. Then the eigenvalues of $H$ and $Z$ interlace,
\begin{equation}
\lambda_1 \leq \mu_1 \leq \lambda_2 \leq \mu_2 \leq..\leq \lambda_n \leq \mu_n.
\nonumber
\end{equation}
\end{corollary}

With these preliminaries, we then prove Theorem ~\ref{theorem:2}. 

\begin{proof}
Let $L$ be a Laplacian matrix of graph $G$, without loss of generality, we remove an edge $e_{uv}$ from $G$ in turn to obtain subgraphs $G'$ of $G$. We consider the variation of the adjacency matrix from the subgraph $G'$ to $G$ when removing one edge $e_{uv}$ from $G$.

\begin{equation}
L(G') = \left\{ 
    \begin{aligned}
    &\sum_{i \neq j} a_{ij}, & &l_{ii} \cr 
    &-a_{ij}, & &l_{ij},i \neq j
    \end{aligned}
\right.
\end{equation}

\begin{equation}
\Delta L = L(G'+e) - L(G') = \left\{ 
    \begin{aligned}
    &1, & &l_{uu}, l_{vv} \cr 
    &-1, & &l_{uv}, l_{vu}
    \end{aligned}
\right.
\end{equation}

Let $p$ is an nonzero vector with element $p_u = 1$ and $p_v = -1$, then $\Delta L = pp^\ast$. There is $\lambda_i(L(G')) \leq \lambda_i(L(G'+e)) = \lambda_i(L(G))$ based on the Corollary \ref{corollary:1}. Similarly, all subgraphs of $G$ can be obtained in this way so that the same conclusions can be reached by just superimposing $pp^\ast$. Consequently, the eigenvalues of the Laplacian matrices of all subgraphs of $G$ are correspondingly smaller than the eigenvalues of the Laplacian matrix of $G$.
\end{proof}

\subsection{Proof of Theorem \ref{theorem:3}}
\label{appendix:A3}
\ 
\begin{proof}
Let $\omega(\hat{S}) = \sum_{i, j \in V, i < j} \omega_{ij} (\hat{x}_i - \hat{x}_j )^2$ denote a weight obtained by a relaxed solution $\hat{S}=\mathcal{A}_\theta(G)$, $\hat{x} \in [0,1]^n$. We analyze the rounding process of the relaxed solution $\hat{S}$ into the discrete solution $S \in \{0,1\}^n$. Let $\hat{x}_i$ and $x_i$, $i = {1,2,...,n}$ denote their entries respectively. The rounding procedure has no requirement on the order of the rounding sequence, w.l.o.g, suppose we round from index $i = 1$ to $i = n$.

In the deterministic rounding phase, we opt for the rounding method that minimizes the loss per round. Additionally, further rounding is conducted for the cases that do not meet the loss inequality. Then the following inequality holds:

\begin{equation}
\begin{split}
\mathcal L(\theta; G) = & \alpha \cdot f(\hat{S};G) + \beta \cdot g(\hat{S};G)\\
= & \alpha \cdot \sum_{i, j \in V, i < j} \omega_{ij} (\hat{x}_i - \hat{x}_j )^2 + \beta \cdot e^{-\tau \cdot [\lambda_3(\hat{S})-\lambda_2(\hat{S})]}\\
\overset{\text{(a)}}{\geq} & \alpha \cdot f([x_1,x_2,...,x_n];G) + \beta \cdot g([x_1,x_2,...,x_n];G)\\
= & \alpha \cdot f(S;G) + \beta \cdot g(S;G)\\
\end{split} 
\end{equation}

\noindent where, (a) denotes the phase of deterministic rounding.
\end{proof}

\underline{\emph{Note.}} We state that finding an entry-wise concave principle similar to \cite{wang2022unsupervised} for the maximum minimal cut problem is challenging.

We consider the case of unweighted graphs for convenience, i.e., $\omega_{ij} = 1$. We begin with the change in cost function $f(\hat{S};G)$ after rounding $\hat{S}$ to $S$.

\begin{equation}
\begin{split}
\omega(\hat{S}) = & \sum_{i, j \in V, i < j} \omega_{ij} \cdot (\hat{x}_i-\hat{x}_j)^2 \\
= & n \sum \hat{x}_i^2 - (\sum \hat{x}_i)^2 \\
= & n (\hat{x}_1^2 + ... +\hat{x}_n^2) - (\hat{x}_1 + ... + \hat{x}_n)^2 
\end{split} 
\end{equation}

Rounding $\hat{x}_1$ to $x_1$ and the difference of them is similarly denoted to $\Delta x_1 = x_1 - \hat{x}_1$. If $\Delta x_i > 0$ then show that $\hat{x}_i$ rounds to $x_i = 1$; $\Delta x_i < 0$ then $\hat{x}_i$ rounds to $x_i = 0$; otherwise $\hat{x}_i$ is unchanged, i.e. $x = \hat{x}_i$.

\begin{equation}
\begin{split}
\omega(\hat{S}+\Delta x_1) = & n[(\hat{x}_1+\Delta x_1)^2 + \hat{x}_2^2 + ... +\hat{x}_n^2] \\
& - (\hat{x}_1+\Delta x_1 + \hat{x}_2+...+\hat{x}_n)^2\\
\end{split} 
\end{equation}

Consequently, the variation of  $\omega(\hat{S})$ after rounding $\hat{x_1}$ to $x_1$ can be obtained.

\begin{equation}
\begin{split}
\Delta \omega(\hat{S}+\Delta x_1) = & \omega(\hat{S}+\Delta x_1) - \omega(\hat{S})  \\
= & n [(\Delta x_1)^2 +2\hat{x}_1\Delta x_1] -(2\sum \hat{x}_i+\Delta x_1) \Delta x_1 \\
= & (n-1)(\Delta x_1)^2 + 2(n\hat{x}_1-\sum \hat{x}_i)\Delta x_1 \\
= & 2(nx_1-\sum \hat{x}_i)\Delta x_1 - (n+1)(\Delta x_1)^2 \\
\end{split} 
\end{equation}

We analyze $\Delta \omega(\hat{S}+\Delta x_1)$ in conjunction with the quadratic function $f(\chi) = a\chi^2 +b\chi +c$. $\Delta x_1$ is regarded as a quadratic function of $\chi$. Thus, it is easy to see that $\Delta \omega$ is hardly uniformly monotonic. In addition, the monotonicity of the eigenvalues is even more difficult to analyze. 
This reveals that the loss function cannot consistently decrease throughout the entire rounding process, which is also one of the reasons why the previous combinatorial solvers are powerless to deal with the MMCP. However, our approach guarantees $\mathcal L(\theta; G) > \mathcal L(S; G)$ directly by rounding procedure, enabling the Theorem \ref{theorem:3} hold.

\subsection{Proof of Theorem \ref{theorem:4}}
\begin{proof}
Let $K_v=[v_1,v_2,...,v_l] \subseteq V$ denote a feasible solution with weight $\psi$ of $G$. Consider the tree obtained from $K_v$ by adding edges $(v_i,v_j) \in E$, $v_i,v_j \in K_v$ until $K_v$ become to a tree $T_K$. The tree $T_{V\backslash K_v}$ of $V\backslash K_v$ can be constructed in the same way. Set $v_p \in K_v$ and $v_q \in V\backslash K_v$, the edge set $F$ connects the cut $K_v$ and $V\backslash K_v$ according to the definition of maximum minimal cut. We can construct a spanning tree $T$ of $G$ by selecting one edge $e$ from $F$ and connecting it with $T_K$ and $T_{V\backslash K_v}$. Hence, the feasible solution with weight $\psi$ of $G$ can be obtained by disconnecting an edge of $T$.   
\end{proof}

\subsection{Other Related Theorem}
\label{appendix:A5}
\begin{theorem}[Matrix-Tree]
\label{theorem:6}
For an undirected graph $G$ with degree matrix $D$ and adjacency matrix $A$. Define the Laplacian matrix $L(G)=D-A$ of $G$. Then the number of spanning trees of $G$ is equal to any one of the algebraic cofactors of $L(G)$.
\end{theorem}

\begin{corollary}
For an undirected graph $G$, the number of connected components of $G$ is equal to the number of zero eigenvalues of the Laplacian matrix.
\end{corollary}

\section{Details of the Proposed Algorithms}
\label{B:details}

\setcounter{table}{0}   %从零开始编号
%定义编号格式，在数字序号前加字符“B"
\renewcommand{\thetable}{B.\arabic{table}}
\setcounter{figure}{0}
\renewcommand{\thefigure}{B.\arabic{figure}}

\subsection{Inherent Complexity of MMCP}
\label{subsection:inc}
Although max-cut and maximum minimal cut are both NP-hard problems, computing the maximum minimal cut of a graph seems to be harder. Interestingly, the max-cut problem can be optimally solved in polynomial time in planar graphs \cite{Hadlock75}, while MMCP was proved NP-complete even on planar graphs in \cite{Haglin91} that showed the MMC on graphs of clique-width $w$ cannot be solved in time $f(w) \times n^{o(w)}$ unless the Exponential Time Hypothesis (ETH) fails.

\subsection{Details of Heuristic Solver}
\label{B:PIONEER}
\noindent \textbf{The case of disconnecting edges.} In Section \ref{subsection: heuristic}, recall that in Phase \uppercase\expandafter{\romannumeral2} of Heuristic, all the edges of $p_{k+1}$ are broke. The following cases may appear.

\begin{figure}[ht]
\centering
\includegraphics[width=1\columnwidth]{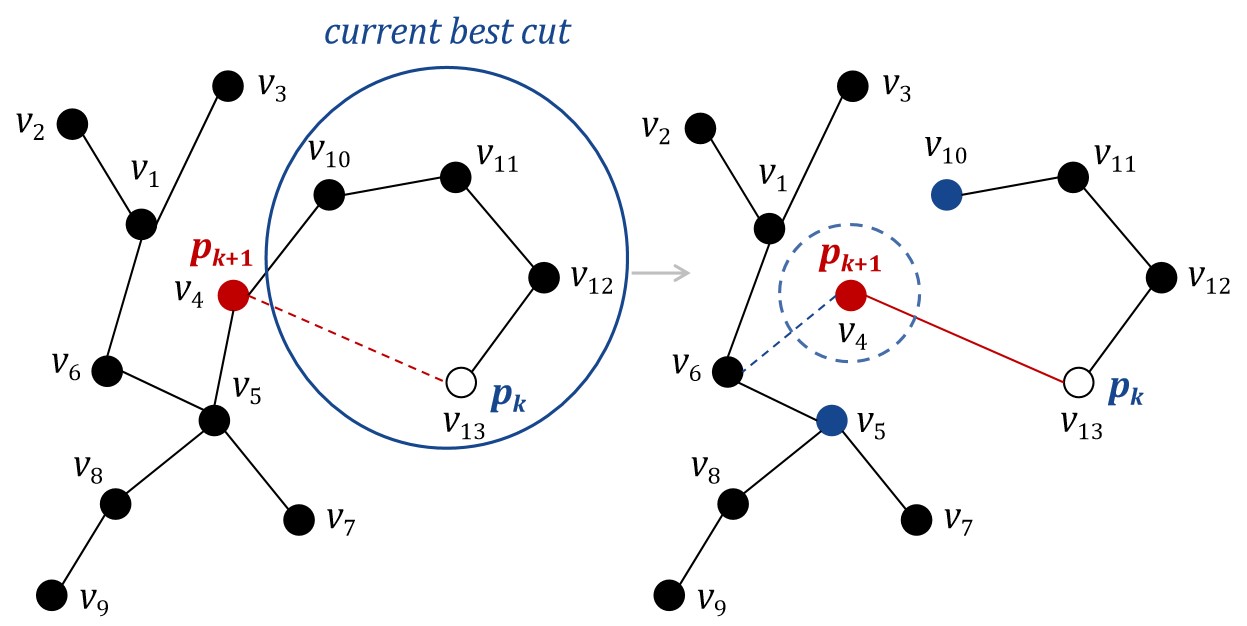}
\caption{An example to demonstrate a special case of $p_{k+1}$.}
\label{fig:B1}
\end{figure}

\begin{figure}[ht]
\centering
\includegraphics[width=1\columnwidth]{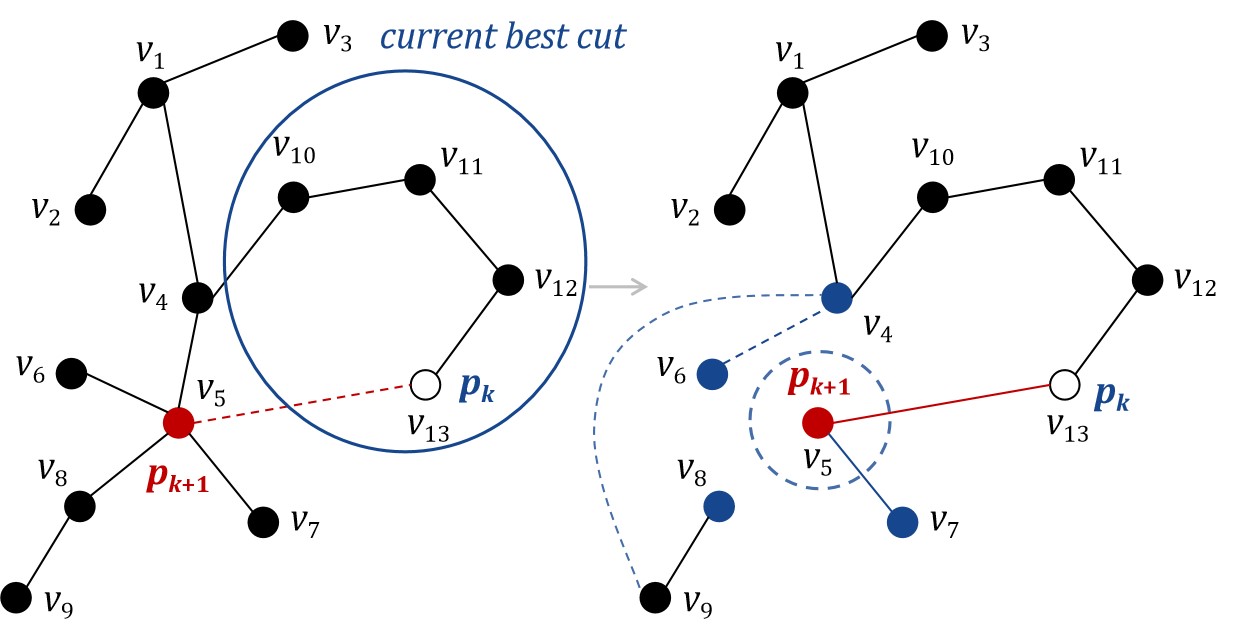}
\caption{An example to indicate a general case of $p_{k+1}$.}
\label{fig:B2}
\end{figure}

\begin{itemize}
\item {\textit{Case}}
1: The current graph is exactly two parts after disconnecting the edges of $p_{k+1}$. Then it only needs to connect one edge between the two connected components to induce a new tree. (e.g. $p_{k+1}=v_4$ in Figure \ref{fig:B1})

\item {\textit{Case}}
2: The disconnected edge will be reserved if the neighbor of $p_{k+1}$ only has neighbor $p_{k+1}$ in $V_2$, and such vertex will be removed from the neighbor of $p_{k+1}$. (e.g. $v_7$ in Figure \ref{fig:B2})

\item {\textit{Case}}
3: The current graph is divided into several connected components after disconnecting the edges of $p_{k+1}$ and it is necessary to connect all the neighbors of $p_{k+1}$ to $V_2$ to induce a new tree. It should be noted, that the corresponding edge should be added according to \emph{case} 1 if $p_{k+1}$ is the disconnected-vertex in $V_1$.
\end{itemize}

\noindent \textbf{The case of adding edges.} In Section \ref{subsection: heuristic} we mentioned that after selecting the vertices in Phase \uppercase\expandafter{\romannumeral3}, we plan to be connected with $p_{k+1}$, the details of adding edges are as follows.

\begin{itemize}
\item {\textit{Case}}
1: The neighbor of $p_{k+1}$ and $p_{k+1}$ are in the same connected component, so there is no need to add an edge. (e.g. $v_4$ in Figure \ref{fig:B2})

\item {\textit{Case}}
2: The neighbor $n$ of $p_{k+1}$ has neighbor $sn$ belong to $V_2$ and the connected component contained $p_{k+1}$, then it is necessary to select a vertex in $sn$, the vertex satisfies the condition, and connect such a vertex to $n$. (e.g. $v_6$ in Figure \ref{fig:B2})

\item {\textit{Case}}
3: The neighbor $n$ of $p_{k+1}$ does not meet the above cases, then we select the vertex $n_c$ in the connected component contained $n$ who has neighbors $sn$ belongs to $V_2$ or another connected component. The edge between such vertex and $n_c$ will be added.(e.g. $v_8$ in Figure \ref{fig:B2})
\end{itemize}

However, we find that the process of adding vertices is time-consuming. Therefore, we opt to construct a spanning tree as a substitute, thereby improving the efficiency of the algorithm.

\begin{figure*}[ht]
\centering
\includegraphics[width=1\textwidth]{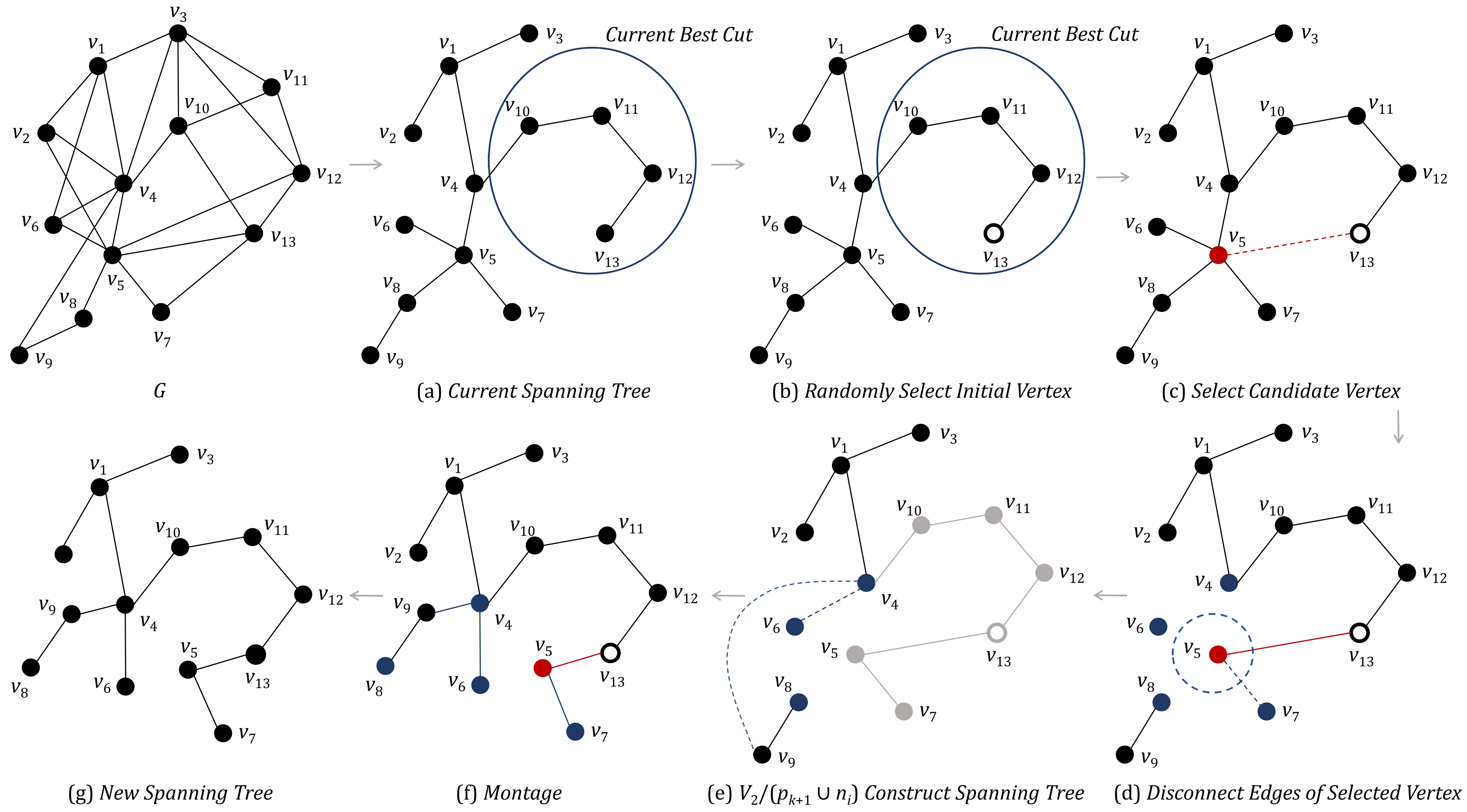}
\caption{Process of adding vertices to build a new spanning tree.}
\label{fig:B3}
\end{figure*}

\noindent \textbf{The pipeline of heuristic solver.} The heuristic solver initially selects a starting vertex based on the optimal cut induced by the current spanning tree. It identifies a candidate set by choosing vertices connected to the selected vertex and presented in the target cut. Subsequently, through a sequence of disconnecting edges, adding edges, and splicing operations, a new spanning tree is obtained. The optimal solution on this new spanning tree is then calculated. This process iterates until adding a vertex to the current optimal cut no longer increases the objective value. Figure \ref{fig:B3} shows the full process of adding vertex from one cut to another cut, which brings in spanning tree transformation. To prevent redundant vertices, a final round of vertex deletion is performed to explore the possibility of achieving a more optimal solution. 
%We can facilitate the transformation of the spanning tree using this approach to uncover more effective solutions.

\textbf{Complexity analysis.} Suppose the number of vertices and edges are $n$ and $m$, respectively. The time complexity of constructing a spanning tree is $O(m\log n)$. Determining the candidate set when adding vertices requires $O(n + m)$. The time used to get the maximum solution of the spanning tree is $O(n-1)$. Thus, the time complexity of the heuristic solver is $O(n^3 + n^2 \cdot m\log n)$.

% \begin{algorithm}[!ht]
% \caption{Search maximum minimal cut by PIONEER.}
% \label{alg:alg1}
% \SetKwData{Or}{\textbf{or}}
% \DontPrintSemicolon
% %\SetAlgoLined
% \KwIn {A weighted undirected graph: $G = (V, E, W)$, the initialized unsupervised model $\mathcal{A}_\theta(\cdot)$, learning rate $r$, batch size $b$, the maximum number of iterations: $MaxIter$}
% \KwOut {A set of maximum minimal cut $BestCut$ and its cut value $BestValue$}
%   \Begin{
%   % $aliasNames \gets []$\;
%   \If{$G$ is not connected}{Exit()\;} 
%   Find all bridges of $G$ by Trajan $\gets bridges, w_{b}$ \;
%   Record the current best solution $\gets BestCut, BestValue$ \;
%   Remove $bridges$ from $G$ $\gets$ $G'' = G_1,...,G_t$ \;
%   \For{$G_i \in G''$}{
%     \If{$name_1 \neq name_2$}{
%         $poly_1 \gets Polygon(poly_1)$ \;
%         $poly_2 \gets Polygon(poly_2)$ \;
%         \If{$poly_1 \cap poly_2 \neq \phi$}{
%           $CurrentValue = $ the cut value of the graph $G$ between $Nodes$ and $OtherNodes.$ \;
%           \If{$CurrentValue > BestValue$}{
%           $BestValue = CurrentValue$ \;
%           $BestCut = [Nodes, OtherNodes]$ \;
%           }
%         }
%     }
%   }
%   \Return{$BestCut, BestValue$}}
% \end{algorithm}

\subsection{Exact Search Algorithm}
\label{appendix:B3}
To verify the accuracy of the proposed method on small graphs, we developed an algorithm for brute force search to achieve exact solutions. The graph is first simplified by graph partitioning. We traverse all possible vertex subsets, checking whether they form valid cuts, and return the vertex subset with the maximum cut value along with its value. We give the pseudo-code of in Algorithm \ref{alg:alg1}.

\textbf{Complexity analysis.} The time complexity of the brute force search algorithm is $O(2^n)$, where $n$ is the number of vertices. It is not applicable for solving MMCP on larger-scale graphs.

\subsection{Random Search Algorithm}
\label{appendix:B4}

\begin{figure}[h]
\centering
\includegraphics[width=1\columnwidth]{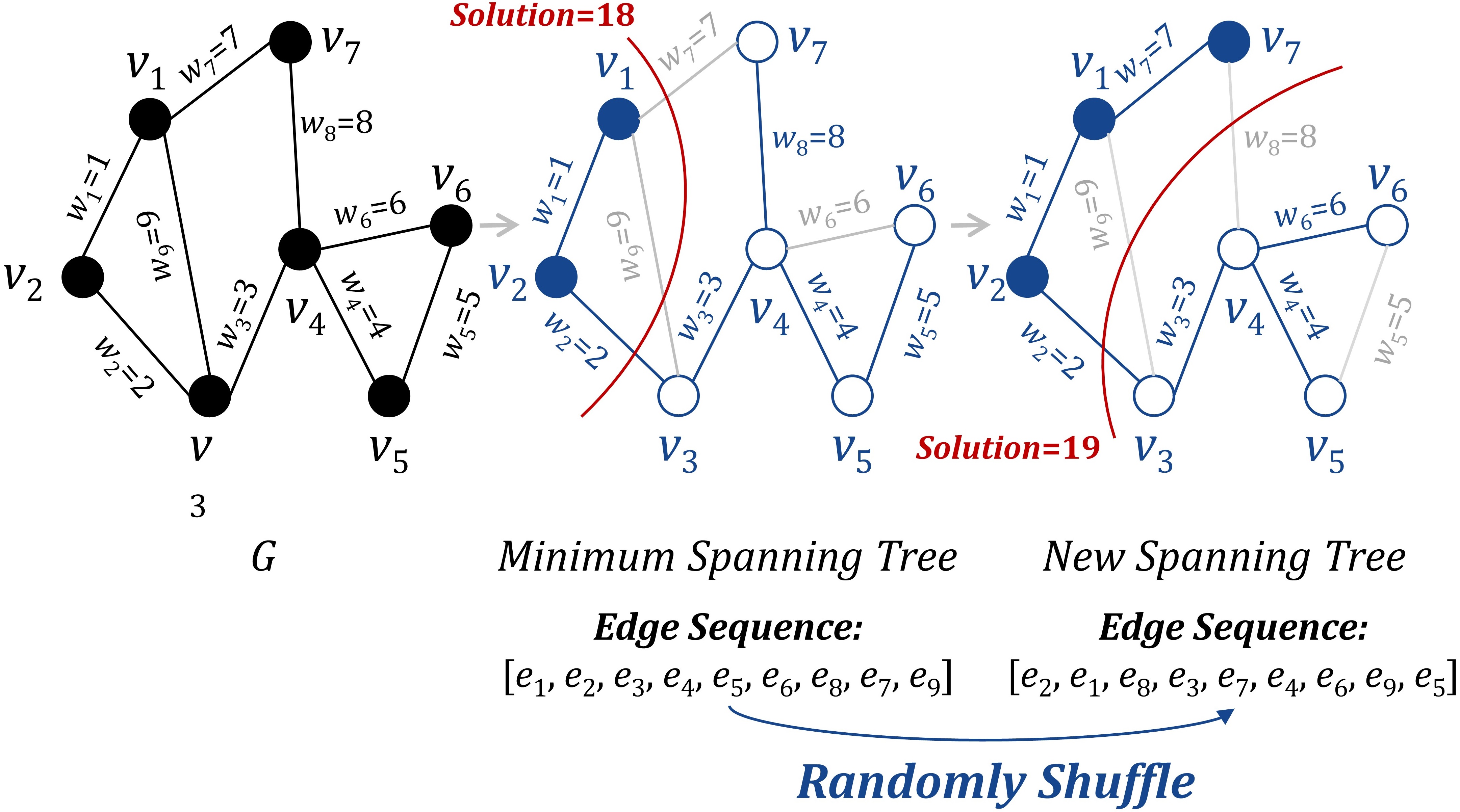}
\caption{An instance to show that randomly shuffle edge sequence to construct a new spanning tree.}
\label{fig:B4}
\end{figure}

We designed a random tree search algorithm in Section \ref{subsection:rts}. The demonstration and the pseudo-code of the algorithm are given in Figure \ref{fig:B4} and Algorithm \ref{alg:alg2}, respectively. 

\begin{algorithm}[!ht]
\caption{Brute force search algorithm.}
\label{alg:alg1}
\SetKwData{Or}{\textbf{or}}
\DontPrintSemicolon
%\SetAlgoLined
\KwIn {A weighted undirected graph: $G = (V, E, W)$}
\KwOut {A set of maximum minimal cut $BestCut$ and its cut value $BestValue$}
\Begin{
% $aliasNames \gets []$\;
  \If{$G$ is not connected}{Exit()\;} 
  Find all bridges of $G$ by Trajan $\gets bridges, w_{b}$ \;
  Record the current best solution $\gets BestCut, BestValue$ \;
  Remove $bridges$ from $G$ $\gets$ $G' = G_1,...,G_t$ \;
  \For{$G_i \in G'$}{
    \For{$cut$ in subset($G.nodes$)}{
        \If{bool($cut$) and len($cut$) $\leq$ len($G_i.nodes$) $/ 2$}{
        $\overline{cut} = G.nodes - cut$ \;
          \If{$G.subgraph(cut)$ and $G.subgraph(\overline{cut})$ are both connected}{
            Compute the cut value of the graph $G_i$ between $cut$ and $\overline{cut} \gets Value$ \;
                \If{$Value > BestValue$}{
                    Update $BestValue$ and $BestCut $ \;
          }
        }
      }
    }
  }
  \Return{$BestCut, BestValue$}}
\end{algorithm}

\begin{algorithm}[!ht]
\caption{Random tree search algorithm.}
\label{alg:alg2}
\SetKwData{Or}{\textbf{or}}
\DontPrintSemicolon
%\SetAlgoLined
\KwIn {A weighted undirected graph: $G = (V, E, W)$, the maximum number of iterations $MaxIter$}
\KwOut {A set of maximum minimal cut $BestCut$ and its cut value $BestValue$}
  \Begin{
  % $aliasNames \gets []$\;
  \If{$G$ is not connected}{Exit()\;} 
  Find all bridges of $G$ by Trajan $\gets bridges, w_{b}$ \;
  Record the current best solution $\gets BestCut, BestValue$ \; 
  Remove $bridges$ from $G$ $\gets$ $G' = G_1,...,G_t$ \;
  \For{$G_i \in G'$}{
    \For{$j$ from $1$ to $MaxIter$}       {Randomly shuffle edge sequence \;
      Construct spanning tree of $G_i$ by Kruskal according to the edge sequence $\gets Tree$ \;
      \For{$edge$ in $Tree$}{
        Break off $edge$ of $Tree \gets Cut, Value$ \;
          \If{$Value > BestValue$}{
            Update $BestValue$ and $BestCut $ \;}
      }
    }
  }
  \Return{$BestCut, BestValue$}}
\end{algorithm}

\textbf{Complexity analysis.} The time complexity of the random tree search algorithm is $O(m\log n + n - 1)$, where $n$ and $m$ are the number of vertices and edges, individually. It is not suitable for solving MMCP of graphs that differ too much from the tree structure.

\subsection{Genetic Algorithm}
\label{appendix:B5}
To further assess the performance of the proposed heuristic solver, we adaptively modified a classical heuristic, Genetic Algorithm (GA) \cite{benabbou2020interactive}, as a baseline. The graph is initially simplified through graph partitioning. Linear ranking selection is employed to generate the parent population. After obtaining new individuals through crossover and mutation, the solutions are evaluated for bi-connectivity requirements using Depth-First Search (DFS). The fitness of a valid solution is defined by its cut value, while solutions that do not meet the constraint are assigned a fitness of $0$. The pseudo-code of the genetic algorithm for the MMCP is as shown in Algorithm \ref{alg:alg3}.

\begin{algorithm}[!ht]
\caption{Genetic algorithm.}
\label{alg:alg3}
\SetKwData{Or}{\textbf{or}}
\DontPrintSemicolon
%\SetAlgoLined
\KwIn {A weighted undirected graph: $G = (V, E,W)$, the maximum number of iterations $MaxIter$, population size $MaxSize$}
\KwOut {A set of maximum minimal cut $BestCut$, its cut value $BestValue$, and its legality label $IsValid$}
  \Begin{
  % $aliasNames \gets []$\;
  \If{$G$ is not connected}{Exit()\;} 
  Find all bridges of $G$ by Trajan $\gets bridges, w_{b}$ \;
  Record the current best solution $\gets BestCut, BestValue$ \; 
  Remove $bridges$ from $G$ $\gets$ $G' = G_1,...,G_t$ \;
  \For{$G_i \in G'$}{
    Randomly generate an initial chromosome \;
    $MaxSize -= 1$ \;
    Construct random initial population with $MaxSize$  \;
    Compute the fitness of each chromosome \;
    \For{$j$ from $1$ to $MaxIter$}{
      \For{$k$ from $1$ to $MaxSize$}{
        Select a pair of chromosomes from population according to fitness \;
        Crossover operation of the selected pair \;
        Mutation of the offspring \;
        Compute the new fitness based on DFS \;
        Add the new chromosome to a new population \;}
      Update population \;
      \If{$Value > BestValue$}{
        Update $BestValue$ and $BestCut $ \;}
      }
      \If{$BestValue = 0$}{
        $IsValid = 0$ \;
        $BestCut = None$ \;}
      \Else{
        $IsValid = 1$ \;}
    }
  }
  \Return{$BestCut, BestValue, IsValid$}
\end{algorithm}

\setcounter{table}{0}   %从零开始编号
%定义编号格式，在数字序号前加字符“C"
\renewcommand{\thetable}{C.\arabic{table}}
\setcounter{figure}{0}
\renewcommand{\thefigure}{C.\arabic{figure}}

\section{More Details of the Experiments}

\subsection{Datasets}
\label{appendix:C1}

\noindent \textbf{Synthetic Mnist datasets.} 
Synthetic Mnist \footnote{\url{http://yann.lecun.com/exdb/mnist/}} datasets are generated according to Section \ref{subsection:mr}. In summary, the synthetic Mnist datasets are classified into two primary types as explained below.

(1) \emph{Isomorphic graph.} The datasets are constructed by individually setting the number of vertices and edges to be 36 and $m$, where $m = 60, 120, 180, 240, 300$. These graphs are undirected, connected, and do not contain bridges, encoded as `I-36|$m$'.

(2) \emph{Heterogeneous graph.} Undirected connected graphs without bridges are built whose number of vertices is $36$,  named `H-36'.

We divide all the datasets into training, validating, and testing sets and count all the constructed datasets as shown in Table \ref{tab:sm}. It is important to note that all synthetic graphs used are undirected, weighted, and connected graphs without bridges.

\begin{table}[H]
  \caption{Statistics of the synthetic Mnist datasets. We give a range of values for the datasets with different numbers of edges.}
  \label{tab:sm}
  \centering
  \begin{tabular}{ccccc}
  \hline
  Dataset & Task & \# Data & \#Vertex & \#Edge \\
  \hline
  \multirow{3}{*}{\makecell{I-36|$m$}} & Train & 8000 & \multirow{6}{*}{36} & \multirow{3}{*}{\makecell{60/120/180/240/300}} \\
  ~ & Valid & 1000 & ~ & ~ \\
  ~ & Test & 1000 & ~ & ~ \\
  \multirow{3}{*}{\makecell{H-36}} & Train & 8000 & ~ & [265, 359] \\
  ~ & Valid & 1000 & ~ & [278, 359] \\
  ~ & Test & 1000 & ~ & [278, 350] \\ \hline
\end{tabular}
\end{table}

\noindent \textbf{Real-world datasets.} There are many real-world datasets commonly used for combinatorial optimization problems on graphs. However, there are no labeled datasets specifically tailored for MMCP, we have selected some typical real-world datasets from these different tasks, which are most undirected, unweighted, and connected graphs. The unconnected graphs are eliminated from all datasets. All datasets are described as follows.

(1) \underline{\emph{ENZYMES dataset.}} The ENZYMES dataset\footnote{\url{https://paperswithcode.com/dataset/enzymes}} is a collection of graph data constructed by the protein tertiary structures obtained from the BRENDA enzyme database \cite{zhao2018work}.

(2) \underline{\emph{IMDB dataset.}} The IMDB-BINARY dataset\footnote{\url{http://www.graphlearning.io/}} is the movie collaboration dataset that is graphs with vertices referring to actors. There will be an edge if two actors appear in the same movie \cite{KKMMN2016}.

(3) \underline{\emph{REDDIT dataset.}} The REDDIT-BINARY dataset \cite{yanardag2015deep} is constructed from online discussion threads in Reddit\footnote{\url{https://www.reddit.com/}} where vertices represent users and edges mean at least one of two users responded to the other user's
comment.

(4) \underline{\emph{Power Grid.}} There are five main power grid datasets, which are coded as 36-vertices, Area 1, Area 2, IEEE118, and IEEE300. Each of these datasets has only one graph and corresponding parameters, where the parameters of IEEE118 and IEEE300 are derived from the results of built-in power flow calculation by PyPower\footnote{\url{https://github.com/rwl/PYPOWER}}. The edge weights signify the average active power on lines. 

The training, validating, and testing sets of the above datasets are composed of subgraphs with $n > 16$ obtained by removing all bridges, self-loops, and duplicate edges of the corresponding original graphs. Thus, we show the statistics of each dataset in Table \ref{tab:rwd}. It is worth mentioning that graphs with a vertex count of $1$ are not included in the subgraphs counted. In addition, we have undertaken case studies on actual power grids, which are categorized as a real-world dataset as well.

\subsection{More Experimental Results}
\label{appendix:C2}
Due to space constraints in the main text, we show more experimental results in this section. It is important to note that all experiments still adhere to the setup outlined in Section \ref{subsection:setup}.

\subsubsection{Supplemental Ablation Study}
\label{appendix:sas}
To further assess the efficacy of the key components in PIONEER, we perform more ablation studies in detail here, including the complete results of Figure \ref{fig:5}, the removal of the graph partitioning from the proposed framework, and the analysis of rounding.

\noindent \textbf{Full results of w/o unsupervised solver.} In table \ref{tab:ar}, we give detailed numerical results in Figure \ref{fig:5} and further report experimental results on real-world datasets. 

\begin{table}[h]
  \caption{Effectiveness of unsupervised solver. The best results are highlighted in bold and execution time is measured in sec. per subgraph (s/g).}
  \label{tab:ar}
  \centering
  \begin{tabular}{ccc}
    \hline
    Dataset & Heuristic & \textbf{PIONEER} \\
    \hline
    I-36|60 & \makecell{\textbf{873.49 $\pm$ 0.65 (0.94 s/g)}}& \makecell{872.34 $\pm$ 0.76 (0.95 s/g)} \\
    I-36|120 & \makecell{\textbf{2391.00 $\pm$ 4.56} (4.70 s/g)}& \makecell{2374.61 $\pm$ 2.43 \textbf{(3.19 s/g)}} \\
    I-36|180 & \makecell{\textbf{3552.04 $\pm$ 1.37} (8.95 s/g)}& \makecell{3543.06 $\pm$ 3.22 \textbf{(5.76 s/g)}} \\
    I-36|240 & \makecell{4551.25 $\pm$ 5.77 (12.15 s/g)}& \makecell{\textbf{4573.49 $\pm$ 4.17} \textbf{(6.46s/g)}} \\
    I-36|300 & \makecell{5496.66 $\pm$ 3.05 (14.69 s/g)}& \makecell{\textbf{5514.47 $\pm$ 4.79} \textbf{(8.45s/g)}} \\
    \makecell{H-36} & \makecell{5638.41 $\pm$ 6.46 (15.29 s/g)}& \makecell{\textbf{5659.22 $\pm$ 6.43} \textbf{(8.46 s/g)}} \\
    \hline
    ENZYMES & \makecell{31.25 $\pm$ 0.63 (2.52 s/g)}& \makecell{\textbf{31.32 $\pm$ 0.43 (2.46s/g)}} \\
    IMDB & \makecell{56.67 $\pm$ 0.12 (2.65 s/g)} & \makecell{\textbf{56.70 $\pm$ 0.10 (1.08 s/g)}}\\
    REDDIT & \makecell{17.71 $\pm$ 0.09 (1.68s/g)} & \makecell{\textbf{17.97 $\pm$ 0.02} \textbf{(1.58s/g)}}\\
    \hline
\end{tabular}
\end{table}

As shown in Table \ref{tab:ar}, the heuristic solver exhibits strong performance on its own, especially on sparse graphs. Nevertheless, our proposed framework presents notable time advantages and attains comparable or even superior quality of solutions. This is attributed to the unsupervised solver providing an improved initial solution for the heuristic, enabling it to more efficiently and swiftly discover the optimal solution. Therefore, we can conclude that unsupervised learning plays a crucial role in accelerating our framework.

\noindent \textbf{w/o Graph partitioning (GP).} The introduction of graph partitioning is aimed at simplifying the scale of the graph, thereby reducing computational complexity to achieve acceleration, the theory of which has been guaranteed by Theorem \ref{theorem:1}. Nevertheless, we execute random algorithms on real-world datasets to more intuitively illustrate the time speedup of graph partitioning. 

\begin{figure}[ht]
\centering
\includegraphics[width=0.85\columnwidth]{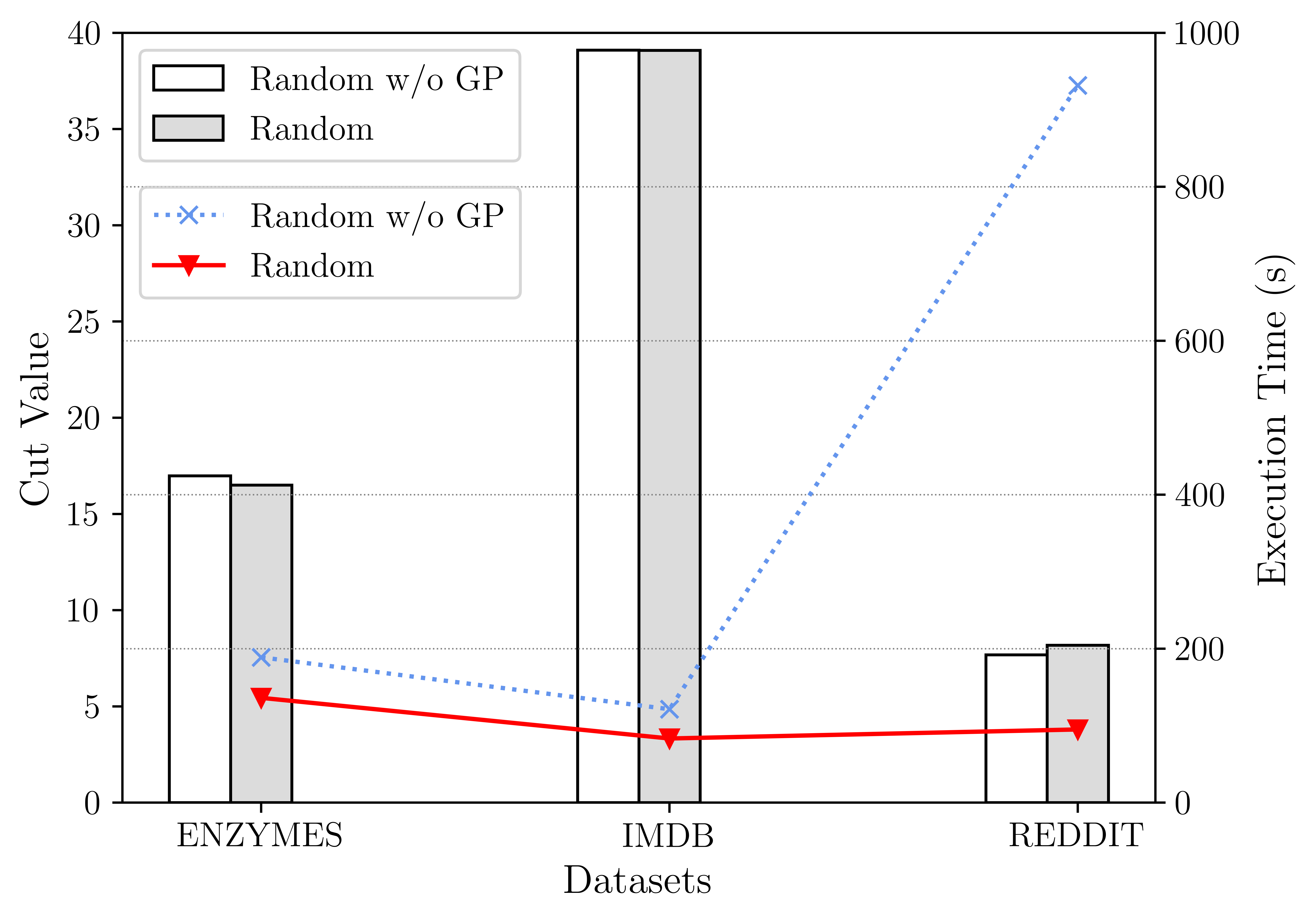}
\caption{Effectiveness of the graph partitioning. We report the results on the real-world datasets. The bars denote the mean of cut values, and the lines denote the execution time.}
\label{fig:C1}
\end{figure}

In Figure \ref{fig:C1}, we observe that the running time employed in graph partitioning is significantly shorter than that of directly solving the original graph. This is attributed to the reduction in graph scale, which effectively lowers complexity and enhances runtime speed for various graph structures. Moreover, it implies that our graph partitioning can serve as an effective starting point for accelerating other MMCP algorithms.

\begin{table*}[h]
  \caption{Effectiveness of the deterministic rounding. The best results are highlighted in bold. The cut values are only reported by mean values for correctly solved instances.}
  \label{tab:round}
  \centering
  \begin{tabular}{ccccccccc}
    \hline
    \multirow{2}{*}{Dataset} & \multicolumn{2}{c}{Round to Nearest Integer} & \multicolumn{2}{c}{Only Phase \uppercase\expandafter{\romannumeral2}} & \multicolumn{2}{c}{Only Phase \uppercase\expandafter{\romannumeral1}} & \multicolumn{2}{c}{\textbf{Unsupervised (Ours)}}\\
    ~ & Violation & Cut Value & Violation & Cut Value & Violation & Cut Value & Violation & Cut Value \\
        \hline
    I-36|60 & 20.58 \% & \makecell{197.78\\(0.01 s/g)} & 36.24 \% & \makecell{187.28\\(0.09 s/g)}  & 10.94 \% & \makecell{580.16\\(0.13 s/g)} & \textbf{4.52} \% & \makecell{551.49\\(0.13 s/g)} \\
    I-36|120 & 14.02 \% & \makecell{1756.47\\(0.02 s/g)} & 0.24 \% & \makecell{1345.72\\(0.10 s/g)} & 3.34 \% & \makecell{2093.41\\(0.13 s/g)} & \textbf{0 \%} & \makecell{2066.12\\(0.12 s/g)} \\
    I-36|180 & 0.48 \% & \makecell{2379.58\\(0.02 s/g)} & 0.24 \% & \makecell{2079.59\\(0.10 s/g)} & \textbf{0 \%} & \makecell{3363.73\\(0.41 s/g)} & \textbf{0 \%} & \makecell{3363.73\\(0.41 s/g)} \\
    I-36|240 & 94.88 \% & \makecell{3676.77\\(0.02 s/g)} & 0 \% & \makecell{3199.41\\(0.10 s/g)} & \textbf{0 \%} & \makecell{4319.44\\(0.14 s/g)} & \textbf{0 \%} & \makecell{4319.44\\(0.14 s/g)} \\
    I-36|300 & 96.28 \% & \makecell{3489.93\\(0.03 s/g)} & 0.04 \% & \makecell{3949.84\\(0.11 s/g)} & \textbf{0 \%} & \makecell{5222.96\\(0.15 s/g)} & \textbf{0 \%} & \makecell{5222.96\\(0.15 s/g)} \\
    \makecell{H-36} & 98.44 \% & \makecell{3603.00\\(0.03 s/g)} & 0.14 \% & \makecell{4122.81\\(0.11 s/g)} & \textbf{0 \%} & \makecell{5372.27\\(0.16 s/g)} & \textbf{0 \%} & \makecell{5372.27\\(0.16 s/g)} \\
    \hline
    ENZYMES & 100 \% & - (0.02 s/g) & 100 \% & - (0.12 s/g) & \textbf{0} \% & 12.62 (0.17 s/g) & \textbf{0} \% & 12.62 (0.17 s/g) \\
    IMDB & 13.47 \% & 9.77 (0.03 s/g) & 22.45 \% & 26.06 (0.06 s/g) & \textbf{4.90} \% & 41.99 (0.07 s/g) & \textbf{4.90} \% & 41.99 (0.07 s/g) \\
    REDDIT & 16.94 \% & 2.98 (0.02 s/g) & 47.29 \% & 3.71 (0.13 s/g) & 4.00 \% & 7.92 (0.16 s/g) & \textbf{1.65} \% & 8.07 (0.16 s/g) \\
    \hline
\end{tabular}
\end{table*}

\noindent \textbf{The analysis of rounding.} In Section \ref{subsection: unsupervised}, we introduce two rounding methods, deterministic rounding and constraint-prior rounding, primarily used to discretize the relaxed solutions obtained by neural networks. We refer to these as the first-phase and second-phase rounding, respectively. Similar to \cite{wang2022unsupervised}, general unsupervised combinatorial solvers directly utilize the loss function as the basis for rounding, selecting the integer solution with a smaller loss at each rounding step. This aligns with rounding phase \uppercase\expandafter{\romannumeral1} as defined in Definition \ref{def:2}. However, we have discovered that network learning is not straightforward due to connectivity being a special property that cannot be directly represented by solutions. Hence, we have devised the second phase of rounding, named constraint-prior rounding. The phase \uppercase\expandafter{\romannumeral2} of rounding serves the main purpose: adjusting the infeasible solution after the phase \uppercase\expandafter{\romannumeral1} of rounding to meet connectivity requirements. Therefore, we conducted comparative experiments on all datasets. In addition, we used the common rounding method as a baseline, i.e. rounding up relaxed solutions to the nearest integer, which does not satisfy Theorem \ref{theorem:3}. It is important to note that we only demonstrate the impact on unsupervised solvers, as heuristic solver consistently finds satisfactory solutions.

We evaluate the effectiveness of different rounding methods by progressively removing them one by one. The results are presented in Table \ref{tab:round}. Although the method that directly rounds to the nearest integer achieves the fastest rounding speed, it only satisfies constraints at a minimal level. This suggests that obtaining valid solutions solely through simple rounding becomes challenging once there is no assurance from Theorem \ref{theorem:3}. Additionally, while constraint-prior rounding can improve the legality of solutions, its cost cannot be guaranteed and may result in lower but still acceptable cut values. It is also evident that satisfactory results can be obtained solely through deterministic rounding, owing to the performance guarantees provided by Theorem \ref{theorem:3}. In contrast, the complete version achieves the highest objective value, as well as the highest solution legitimacy rate and fast processing speed.

Meanwhile, the results also reveal that learning connectivity for networks might not be as straightforward, especially for connectivity learning in different graph structures. This insight inspired us to propose heuristic algorithms, precisely to enhance and repair the solutions of the unsupervised solver, addressing MMCP in various graph structures.

\noindent \textbf{w/o Upper bound $\lambda_3(G)$.} Recall that the penalty function $g(\hat{S};G)) = e^{-\tau \cdot [\lambda_3(\hat{S})-\lambda_2(\hat{S})]}$ and $\tau = ln\epsilon/\lambda_3(G)$, which has two key parameters, i.e. the adjustment parameter $\epsilon$ and the upper bound $\lambda_3(G)$. Here, we only analyze the case without upper bound $\lambda_3(G)$ and leave the analysis of $\epsilon$ in \ref{subsection:sa}. 

For different graphs, there are noticeable differences in the eigenvalues. The introduction of $\lambda_3(G)$ is aimed at minimizing the impact of eigenvalue disparities on the learning process. As part of the ablation study, we set $\tau = 1$, that is, penalty is $e^{-[\lambda_3(\hat{S})-\lambda_2(\hat{S})]}$ for all graphs. 

\begin{table*}[h]
  \caption{Parameter study of $\epsilon$. The best results are highlighted in bold. The cut values are only reported by mean values for correctly solved instances.}
  \label{tab:epsilon}
  \centering
  \begin{tabular}{ccccccccc}
    \hline
    \multirow{2}{*}{Dataset} & \multicolumn{2}{c}{$\epsilon = 0.1$} & \multicolumn{2}{c}{$\epsilon = 0.01$} & \multicolumn{2}{c}{$\epsilon = 0.001$} & \multicolumn{2}{c}{\textbf{Unsupervised (Ours)}}\\
    ~ & Violation & Cut Value & Violation & Cut Value & Violation & Cut Value & Violation & Cut Value \\
        \hline
    I-36|60 & 2.24 \% & 335.10 & 0.16 \% & 369.32 & 2.04 \% & 459.19 & \textbf{4.52 \%} & \textbf{551.49} \\
    I-36|120 & 1.04 \% & 1544.07 & 1.60 \% & 1550.81 & 0 \% & 2023.97 & \textbf{0 \%} & \textbf{2069.82} \\
    I-36|180 & 0.04 \% & 3292.11 & 0 \% & 3247.04 & 0 \% & 3342.81 & \textbf{0 \%} & \textbf{3363.73} \\
    I-36|240 & 0 \% & 4342.85 & 0 \% & 4303.53 & \textbf{0 \%} & \textbf{4373.97} & 0 \% & 4319.44 \\
    I-36|300 & \textbf{0 \%} & \textbf{5230.20} & 0 \% & 5169.11 & 0 \% & 5187.31 & 0 \% & 5222.96 \\
    \makecell{H-36} & 0 \% & 5355.65 & 0 \% & 5330.92 & 0 \% & 5346.43 & \textbf{0 \%} & \textbf{5372.27} \\
    \hline
    ENZYMES & 97.44 \% & 8.00 & 100 \% & - & 46.15 \% & 12.29 & \textbf{0} \% & \textbf{12.62} \\
    IMDB & 3.67 \% & 25.79 & 6.94 \% & 44.21 & 14.69 \% & 32.05 &\textbf{4.90 \%} & \textbf{41.99}\\
    REDDIT & 25.41 \% & 6.23 & 13.65 \% & 8.61 & 7.06 \% & 6.95 & \textbf{1.65 \%} & \textbf{8.07} \\
    \hline
\end{tabular}
\end{table*}

The results are summarized in Table \ref{tab:lambda}. We can see that the variant without $\lambda_3(G)$ performs the worst in terms of the legitimacy of solutions, which is consistent with our envision. If $\tau$ is fixed, the uniform dimension of the penalty for all graphs would make it challenging to distinguish whether the solutions are valid for graphs of different scales, thereby resulting in a decline in learning performance. This shows the need to combine the upper bound of $\lambda_3$ with the penalty in the learning process of different graphs. 

\begin{table}[h]
  \caption{Effectiveness of $\lambda_3(G)$. The significantly worse results are highlighted in underlined. The cut values are only reported by mean values for correctly solved instances.}
  \label{tab:lambda}
  \centering
  \begin{tabular}{ccccc}
    \hline
    \multirow{2}{*}{Dataset} & \multicolumn{2}{c}{w/o $\lambda_3(G)$} & \multicolumn{2}{c}{\textbf{Unsupervised (Ours)}}\\
    ~ & Violation & Cut Value & Violation & Cut Value  \\
    \hline
    I-36|60 & \underline{88.48 \%} & 549.90 & 4.52 \% & 551.49 \\
    I-36|120 & \underline{1.70 \%} & 1540.14 & 0 \% & 2066.12 \\
    I-36|180 & 0 \% & 3280.54 & 0 \% & 3216.96 \\
    I-36|240 & 0 \% & 4308.04 & 0 \% & 4319.44 \\
    I-36|300 & 0 \% & 5334.66 & 0 \% & 5222.96 \\
    \makecell{H-36} & 0 \% & 5499.91 & 0 \% & 5372.27 \\
    \hline
    ENZYMES & \underline{74.36 \%} & 13.21 & 0 \% & 12.62 \\
    IMDB & 2.04 \% & 32.48 & 4.90 \% & 41.99 \\
    REDDIT & \underline{59.53 \%} & 9.34 & 1.65 \% & 8.07  \\
    \hline
\end{tabular}
\end{table}

\subsubsection{Parameter Study.}
\label{subsection:sa}
We conduct parameter studies to analyze key parameters, consisting of the self-adaptive coefficient $\epsilon$ for penalty and the balanced control parameters $\alpha$ for cost and penalty. We discuss the results in detail as follows.

\noindent \textbf{The parameter study of $\epsilon$.} Recall that we previously analyzed the case without upper bound $\lambda_3(G)$ for the penalty $g(\hat{S};G))$. Next, we analyze the adjustment parameter $\epsilon$ (default $0.0001$). All experiments are performed on all datasets. As a comparison, set $\epsilon = 0.1, 0.01, 0.001$. Since there is no significant difference between the execution time of different $\epsilon$, the running time is not recorded. Similar to the version without $\lambda_3$, $\epsilon$ mainly affects the learning of penalties by the network.

As the result in Table \ref{tab:epsilon} shows, by varying $\epsilon$ from $0.1$ to $0.0001$, we observe that our model achieves the best overall performance among different datasets when $\epsilon$ is around $0.0001$. The results indicate that a smaller value of $\epsilon$ introduces more differentiation in determining the legality of solutions. This is because a larger $\epsilon$ would impose a significant penalty even on solutions that already satisfy the requirements, leading to instability in the learning process. Overall, a
moderate value around $0.0001$ can be a suitable choice for various datasets.

\noindent \textbf{The parameter study of $\alpha$.} Recall that the loss function $\mathcal L(\theta; G)$ $\triangleq \alpha \cdot f(\hat{S};G) + \beta \cdot g(\hat{S};G))$. $\alpha$ is a control parameter that balances cost and penalty to improve the quality of the solution. To illustrate the role of $\alpha$, we conducted experiments on all datasets. We set $\alpha = 1$ and $\alpha = m/n$ for all datasets to determine the appropriate value. For the same considerations as before, only the results of the unsupervised solver are reported. As both solvers exhibit nearly identical runtime, execution times are not presented. 

\begin{table}[h]
  \caption{Parameter study of $\alpha$. The best results are highlighted in bold. The cut values are only reported by mean values for correctly solved instances.}
  \label{tab:alpha}
  \centering
  \begin{tabular}{ccccc}
    \hline
    \multirow{2}{*}{Dataset} & \multicolumn{2}{c}{$\alpha = 1$} & \multicolumn{2}{c}{$\alpha = m/n$}\\
    ~ & Violation & Cut Value & Violation & Cut Value  \\
        \hline
    I-36|60 & 4.68\% & 453.95 & \textbf{4.52 \%} & \textbf{551.49} \\
    I-36|120 & 0 \% & 1762.39 & \textbf{0 \%} & \textbf{2069.82} \\
    I-36|180 & 0 \% & 2754.60 & \textbf{0 \%} & \textbf{3363.73} \\
    I-36|240 & 0 \% & 3491.16 & \textbf{0 \%} & \textbf{4319.44} \\
    I-36|300 & 0 \% & 1956.11 & \textbf{0 \%} & \textbf{5222.96} \\
    \makecell{H-36} & 0 \% & 1264.93 & \textbf{0 \%} & \textbf{5372.27} \\
    \hline
    ENZYMES & 48.72 \% & 10.75 & \textbf{0 \%} & \textbf{12.62} \\
    IMDB & \textbf{4.90} \% & \textbf{41.99} & 31.43\% & 58.91 \\
    REDDIT & \textbf{1.65} \% & \textbf{8.07} & 24.00 \% & 10.77  \\
    \hline
\end{tabular}
\end{table}

As the results reported in Table \ref{tab:alpha}, for graphs with a larger number of edges, $\alpha=m/n$ can effectively explore solutions with larger cut values while ensuring feasibility. This is attributed to the relatively simpler learning of connectivity structures in isomorphic graphs. In contrast, in real-world datasets where graph structures are more special and diverse, fixing $\alpha$ proves more advantageous for prioritizing connectivity constraints, leading to a significant increase in the number of feasible solutions.

\begin{figure*}[ht]
\centering
\includegraphics[width=1\textwidth]{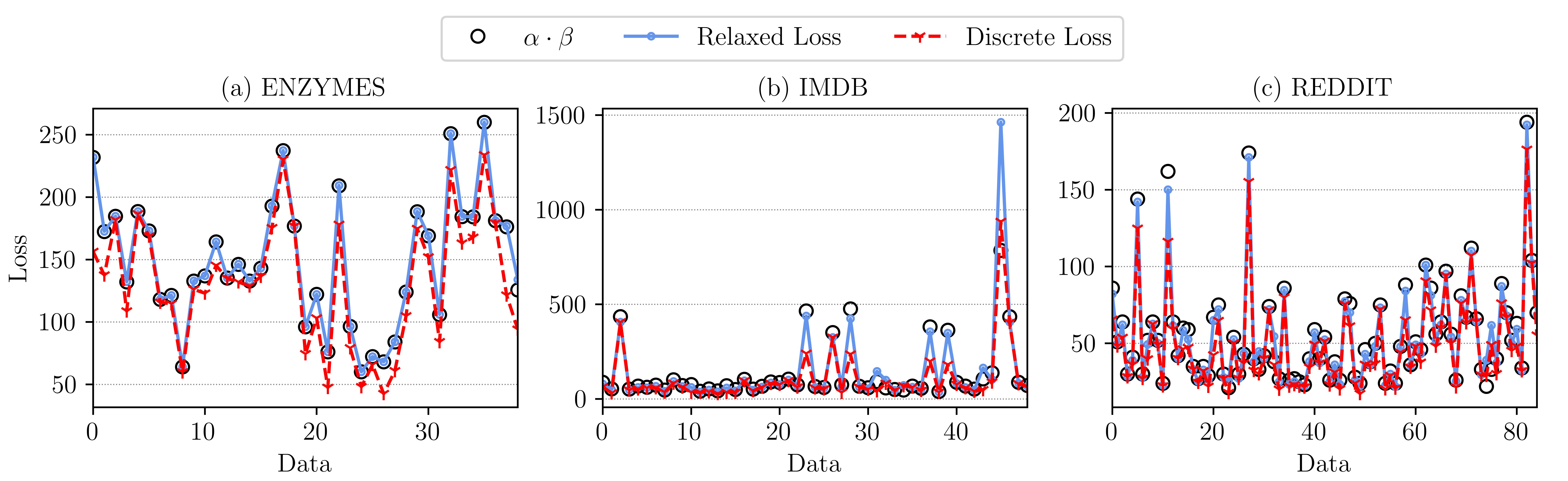}
\caption{Compare the loss functions of relaxed solutions, discrete solutions, and the value of $\alpha \cdot \beta$.}
\label{fig:C2}
\end{figure*}

\subsubsection{Performance Guarantee Study.}
\label{subsection:pgs}

To illustrate the performance guarantee of the unsupervised solver on Theorem \ref{theorem:3}, we compare $\alpha \cdot \beta$, the relaxed loss, and the discrete loss of the training process for real-world datasets in Figure \ref{fig:C2}. According to Theorem \ref{theorem:3}, when the relaxed loss $\mathcal L(\hat{S}; G)$ is learned to be sufficiently small, i.e., $\mathcal L(S; G) < \mathcal L(\hat{S}; G)<\alpha \cdot \beta$, a low-cost and feasible solution can be obtained by deterministic rounding. Therefore, our aim is to ultimately achieve the minimal discrete loss among them.

As shown in Figure \ref{fig:C2}, for discrete solutions with relaxed losses less than $\alpha \cdot \beta$, all discrete solutions have losses lower than $\alpha \cdot \beta$. In other words, the obtained discrete solutions are guaranteed by Theorem \ref{theorem:3}, ensuring both constraint satisfaction and lower costs. Surprisingly, for cases where the relaxation loss is greater than $\alpha \cdot \beta$, the discrete solutions obtained through our proposed rounding not only make $\mathcal L(S; G) < \mathcal L(\hat{S}; G)$ hold but also ensure the validity of $\mathcal L(S; G) <\alpha \cdot \beta$ for the majority of solutions, fulfilling the constraints. Overall, our model can ensure performance without analyzing the entire rounding process for loss fluctuations in addressing MMCP.

\subsubsection{Heuristic Study.}
\label{subsection:hs}
To further demonstrate the competitiveness of the proposed heuristic solver, we adaptively modify the classical genetic algorithm (GA) in Section \ref{appendix:B5} and employ it as a baseline. The maximum number of iterations is set to $100$ for the revised genetic algorithm and we record the optimal solution among them. Table \ref{tab:heuristic} summarizes the results on real-world datasets to contrast different heuristics.

\begin{table}[h]
  \caption{Comparison of different heuristics on real-world datasets. The best results are highlighted in bold. The cut values are only reported by mean values for correctly solved instances.}
  \label{tab:heuristic}
  \centering
  \begin{tabular}{ccccc}
    \hline
    \multirow{2}{*}{Dataset} & \multicolumn{2}{c}{Genetic Algorithm} & \multicolumn{2}{c}{\textbf{Heuristics (Ours)}}\\
    ~ & Violation & Cut Value & Violation & Cut Value  \\
    \hline
    ENZYMES & 57.44 \% & \makecell{27.08\\(6.16 s/g)} & \textbf{0 \%} & \textbf{\makecell{31.25\\(2.52 s/g)}} \\
    IMDB & 27.76 \% & \makecell{71.53\\(36.92 s/g)} & \textbf{0 \%} & \makecell{56.67\\\textbf{(2.65 s/g)}} \\
    REDDIT & 52.00 \% & \makecell{11.59\\(6.52 s/g)} & \textbf{0 \%} & \textbf{\makecell{17.71\\(1.68 s/g)}}  \\
    \hline
\end{tabular}
\end{table}

As shown in Table \ref{tab:heuristic}, our heuristic solver surpasses the competitor in terms of solution legitimacy and running time, which benefits from Theorem \ref{theorem:4}. The results imply that the performance is worse if judging whether the solution satisfies the constraints after its generation because the solution space is not compressed into the feasible domain. We also observe that the proposed heuristic solver achieves larger cut values on datasets other than IMDB, primarily due to the significant variation in the cut values of IMDB solutions. Excluding illegal solutions may result in a higher average cut value compared to the case that all solutions are legal. Overall, our heuristic efficiently explores solutions of higher quality, ensuring adherence to the bi-connected constraint.

\subsubsection{Details of Case Study}
\label{subsection:cs}
The power grid is a vast and intricate network consisting of diverse components that can be represented as a connected graph comprising vertices and edges. The cross-section identification of the power grid is an extremely important and challenging task in power system analysis, which can be formalized mathematically as a maximum minimal cut problem.

\begin{figure}[ht]
\centering
\includegraphics[width=1\columnwidth]{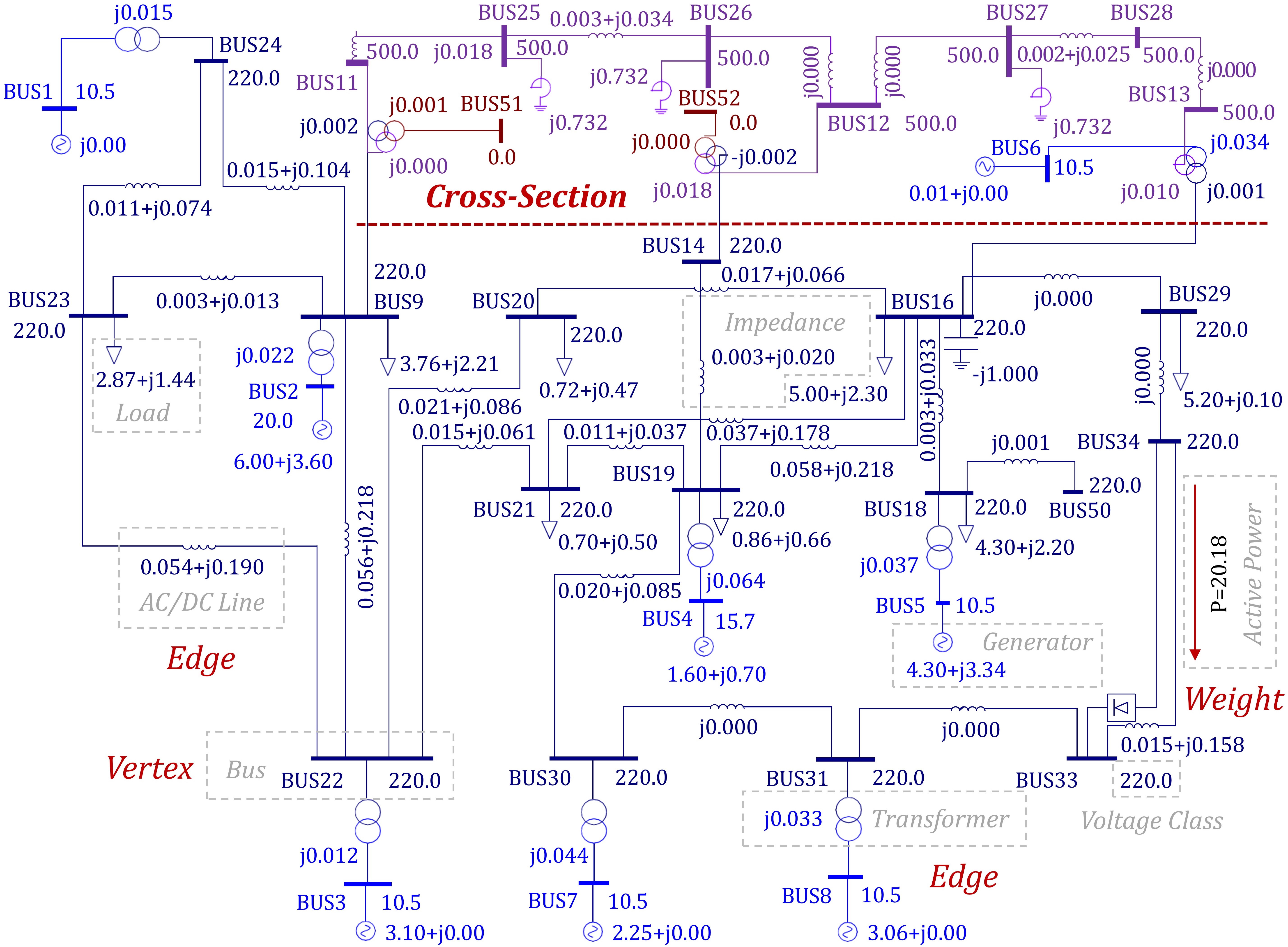}
\caption{An example of a cross-section in power grid. We construct the undirected connected graph by abstracting buses as vertices, AC/DC lines and transformers as edges, and active power flowing on them as the weights of edges.}
\label{fig:C3}
\end{figure}

\noindent \textbf{Problem Statement (Identification of Critical Cross-section).} In the practical power grid, it is necessary to identify the critical cross-section, defined as a problem to find a set of channel sets consisting of several lines with maximum power summation that satisfy certain constraints. Since there is a limit to power delivery from one region to another, the exit or failure of any element in the channel can cause cascading failures, when the system power exceeds the set cross-section value. Therefore, the identification of a critical cross-section of the power grid is crucial to the safe and stable operation of the power system. Moreover, both sides of the critical cross-section must be connected respectively to ensure power supply. Thus, the task of identifying a critical cross-section of the power grid can be abstracted as a maximum minimal cut problem. An example of a cross-section in the power grid is shown in Figure \ref{fig:C3}.

\begin{figure}[h]
\centering
\includegraphics[width=1\columnwidth]{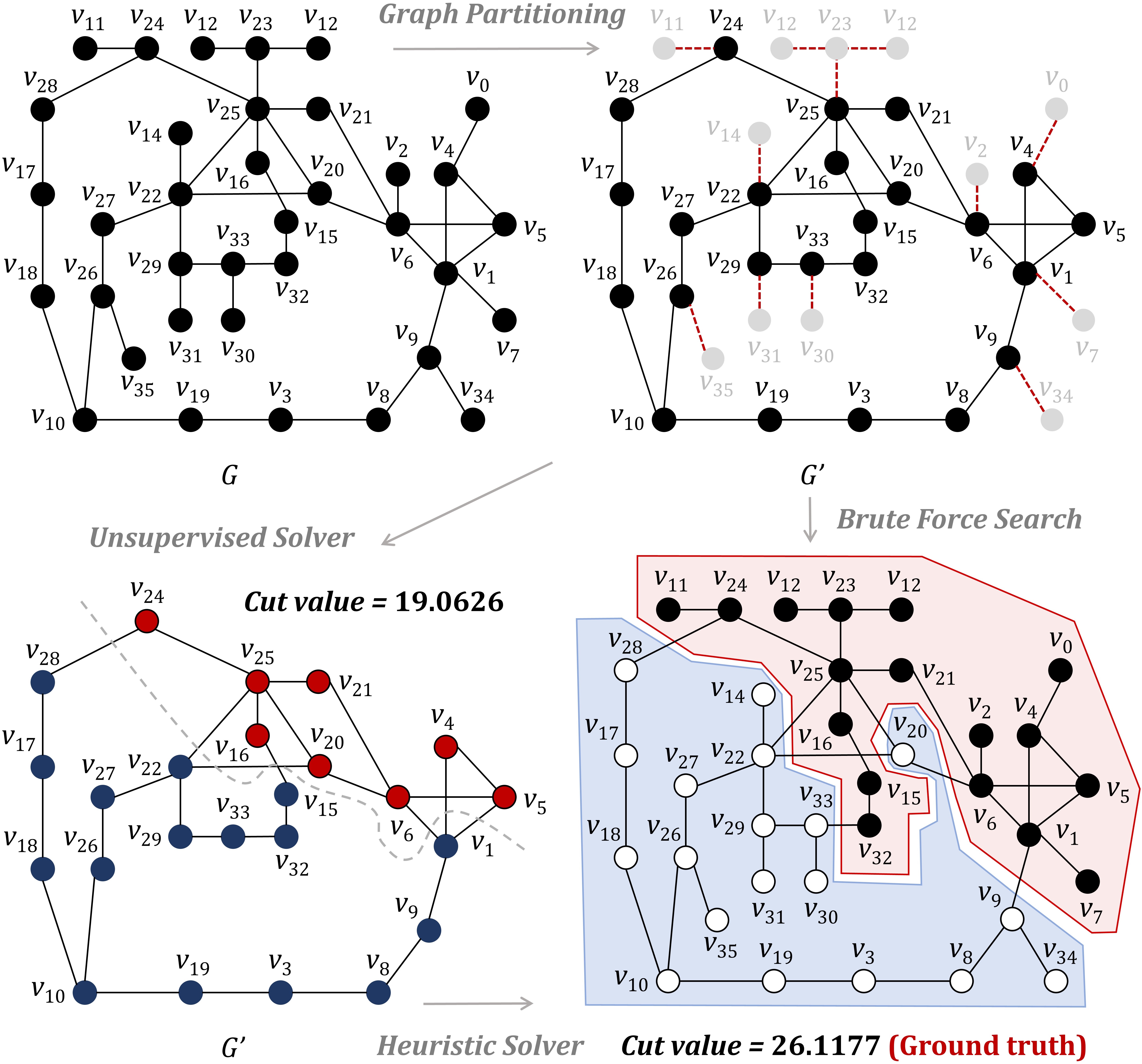}
\caption{Visualize the results on the 36-vertices dataset of the power grid.}
\label{fig:C4}
\end{figure}

\noindent \textbf{Visualization of solutions.} To illustrate how PIONEER works better, we visualize the solution for the 36-vertices of the power grid dataset in Figure \ref{fig:C4}. Graph partitioning first realizes graph simplification and reduces computational complexity. The unsupervised solver achieves a solution similar to ground truth, which is then augmented by a heuristic solver to find the optimal solution. Note that the optimal solution can also be obtained directly by the proposed heuristic solver.

\section{Broader Impact}
In this paper, we introduce an unsupervised framework combined heuristic to resolve the maximum minimal cut problem. The broader impact of this paper is discussed from the following aspects:

1) \emph{What does this study mean for the power grid?} The power grid constitutes a critical component of the global energy system. However, in recent years, there have been recurrent incidents of major power outages. The root cause of these incidents is the occurrence of faults in overloaded transmission lines. When these lines are disconnected, it triggers a transfer of power flows, causing adjacent lines to become overloaded. This, in turn, leads to cascading failures. Therefore, the scope of power grid security analysis can be narrowed down by strategically monitoring critical cross-sections, providing ample time for subsequent load-shedding control strategies. However, in traditional power grid analysis, the identification of critical cross-sections is typically accomplished through complex mechanistic analyses or empirical expertise, lacking rapid and effective methods to address this challenge. Our research provides a significant solution to this difficulty, offering important insights for resolving such issues in power grids.

2) \emph{Who may benefit from this study?} The researchers, engineers, and organizations in the fields of electricity and transportation who employ our framework to address the maximum minimal cut problem may derive significant benefits from this research. This is due to our method's independence from labeled data, coupled with its ability to deliver satisfactory solutions at a low cost and with high quality. On a broader scale, our research can also yield advantages for households and communities, stemming from our exploration of a fundamental problem. This problem holds relevance across diverse network planning and analysis scenarios, thereby supporting subsequent business needs.

3) \emph{Who may be put at risk from this study?} Although our method can ensure the quality of the obtained solution when the loss is low and further enhance the solution quality through a heuristic, the gap between the achieved solution and the true optimal solution remains unclear. Therefore, before deploying our method in scenarios that require strict guarantees on solution approximation, there are still some potential issues that need to be addressed.

\section{Licenses}
We adopt the following datasets in this work, their licenses are listed as follows:

\begin{itemize}
\item The Synthetic Mnist datasets in this work are generated and proposed by us. It is inspired by \cite{poganvcic2019differentiation, wang2022unsupervised} and employs the images from MNIST \cite{deng2012mnist}, which is under the Creative Commons Attribution-Share Alike 3.0 license and is publicly available. Please cite both our paper and their paper in the new publications.
\item The real-world datasets in this work are ENZYMES, IMDB, and REDDIT from \cite{zhao2018work, KKMMN2016, yanardag2015deep} and are publicly available. Please cite their paper in the new publications.
\item The power grid dataset in the case study is from China Electric Power Research Institute which has granted us a license to make it available. Please cite our paper in the new publications.
\end{itemize}

All the datasets and code bases are publicly available except the actual power grid dataset. They contain no human information or offensive content.

\end{document}